\documentclass{article} 
\usepackage[preprint]{neurips_2026}
\usepackage{tikz}
\usepackage{algorithm}

\usepackage{algorithmic}
\usepackage{caption}
\usepackage{multirow}
\usepackage{hyperref}
\usepackage{url}
\usepackage{xcolor}
\usepackage{caption}

\definecolor{algcommentgray}{gray}{0.45}

\newcommand{\ALGComment}[1]{%
  \STATE \textcolor{algcommentgray}{\emph{// #1}}%
}

\DeclareRobustCommand{\figmark}[1]{%
  \tikz[baseline=(X.base)]\node (X)%
    [circle, fill=black, inner sep=1.2pt]%
    {\color{white}\scriptsize\bfseries #1};%
}

\title{RL Excursions during Pre-Training:\\Re-examining Policy Optimization for LLM training}

\author{%
  Rachit Bansal\thanks{Equal contribution. Order decided by a dice roll.}\quad
  Clara Mohri\footnotemark[1]\quad
  Tian Qin\footnotemark[1]\quad
  David Alvarez-Melis\thanks{Equal advising.}\quad
  Sham Kakade\footnotemark[2] \\
  Harvard University \\
  \texttt{\{rachitbansal,cmohri,tqin\}@g.harvard.edu}
}

\newcommand{\cM}{\mathcal{M}}

\usepackage[utf8]{inputenc} 
\usepackage{hyperref}       
\usepackage{url}            
\usepackage{booktabs}       
\usepackage{nicefrac}       
\usepackage{microtype}      
\usepackage{array}
\usepackage{subcaption}   
\usepackage{graphicx}     

\usepackage{wrapfig}

\usepackage{amsmath}
\usepackage{amssymb}
\usepackage{amsthm}
\usepackage{bm,bbm}

\usepackage{listings}
\usepackage[dvipsnames]{xcolor}
\usepackage{cleveref}
\usepackage{mleftright}
\usepackage{enumitem}
\setlist[itemize,1]{leftmargin=1em,itemsep=0.1em,topsep=0.3em}

\usepackage{url}
\usepackage{xparse}
\usepackage{algorithm}
\usepackage{algorithmic}
\usepackage{titlesec}
\titlespacing*{\paragraph}{0pt}{1ex plus .5ex}{0.5ex}
\titleformat{\paragraph}[runin]
  {\normalfont\normalsize\bfseries}
  {\theparagraph}{1em}{}[~~]

\usepackage[bottom,flushmargin]{footmisc}

\usepackage{caption}
\usepackage{subcaption}

\usepackage{natbib}

\usepackage{comment}

\NewDocumentCommand{\loss}{o}{
  \ell 
  \IfValueT{#1}{^{(#1)}}
}
\NewDocumentCommand{\Loss}{o}{
  \gL 
  \IfValueT{#1}{^{(#1)}}
}
\NewDocumentCommand{\lr}{o}{%
  \eta 
  \IfValueT{#1}{^{(#1)}}
}

\NewDocumentCommand{\param}{o}{%
  \theta 
  \IfValueT{#1}{^{(#1)}}
}
\NewDocumentCommand{\Param}{o}{
  \theta 
  \IfValueT{#1}{^{(#1)}}
}
\NewDocumentCommand{\tparam}{o}{%
  \tilde{\theta} 
  \IfValueT{#1}{^{(#1)}}
}
\NewDocumentCommand{\grad}{o}{
  g
  \IfValueT{#1}{^{(#1)}}
}
\NewDocumentCommand{\gradrot}{o}{
  \tilde{g}
  \IfValueT{#1}{^{(#1)}}
}
\NewDocumentCommand{\Grad}{o}{
  G
  \IfValueT{#1}{^{(#1)}}
}

\NewDocumentCommand{\DeltaParam}{o}{
  \Delta
  \IfValueT{#1}{^{(#1)}}%
}
\NewDocumentCommand{\update}{o}{
  \delta
  \IfValueT{#1}{^{(#1)}}%
}
\NewDocumentCommand{\paramCov}{o}{%
  M 
  \IfValueT{#1}{^{(#1)}}
}
\NewDocumentCommand{\tparamCov}{o}{%
  \widetilde{M} 
  \IfValueT{#1}{^{(#1)}}
}

\NewDocumentCommand{\precond}{o}{
  P
  \IfValueT{#1}{^{(#1)}}
}
\NewDocumentCommand{\precondD}{o}{
  D
  \IfValueT{#1}{^{(#1)}}
}

\newcounter{algline} 

\definecolor{RoyalBlue}{RGB}{0,100,170}
\definecolor{peach}{rgb}{1, 0.56, 0.56}
\definecolor{midgray}{RGB}{150,150,150}
\definecolor{EasternBlue}{RGB}{37,150,190}
\definecolor{sand}{RGB}{250,150,120}
\definecolor{grass}{RGB}{120, 190, 50}
\definecolor{sky}{RGB}{50,150,250}
\definecolor{Orange}{RGB}{250,150,50}
\definecolor{Cerulean}{RGB}{80,150,220}
\definecolor{Emerald}{RGB}{62,156,94}
\definecolor{Rouge}{RGB}{250,95,95}
\definecolor{coral}{RGB}{240,128,128}

\definecolor{ColorDef}{RGB}{80, 180, 150}
\definecolor{RevisionRed}{RGB}{240,35,35}
\definecolor{RevisionBlue}{RGB}{80,180,250}
\definecolor{TODO}{RGB}{255,150,50}

\definecolor{shamcolor}{HTML}{1f77b4}   
\definecolor{davidcolor}{HTML}{ffde21}  
\definecolor{depencolor}{HTML}{2ca02c}     
\definecolor{bingbincolor}{HTML}{f26b83}   
\definecolor{rachitcolor}{HTML}{b027d6}  

\newtheorem*{namedtheorem}{\theoremname}
\newcommand{\theoremname}{testing}

\newtheorem*{theorem*}{Theorem}

\newtheorem*{lemma*}{Lemma}

\newtheorem*{corollary*}{Corollary}

\newtheorem*{question*}{Question}

\theoremstyle{definition}

\newtheorem*{definition*}{Definition}

\newtheorem*{remark*}{Remark}

\theoremstyle{plain}

\def\eqref#1{equation~\ref{#1}}

\def\1{\bm{1}}

\DeclareMathAlphabet{\mathsfit}{\encodingdefault}{\sfdefault}{m}{sl}
\SetMathAlphabet{\mathsfit}{bold}{\encodingdefault}{\sfdefault}{bx}{n}

\def\gL{{\mathcal{L}}}

\def\showcomments{} 

\ifdefined\showcomments
  \newcommand{\cm}[1]{\textcolor{red}{CM: #1}}
  \newcommand{\sq}[1]{\textcolor{purple}{SQ: #1}}
  \newcommand{\rachit}[1]{\textcolor{teal}{RB: #1}}
  
   \newcommand{\todo}[1]{\textcolor{pink}{[TODO: #1]}}
\else
  \newcommand{\cm}[1]{}
  \newcommand{\sq}[1]{}
  \newcommand{\rachit}[1]{}
  \newcommand{\todo}[1]{}
\fi

\hypersetup{
  breaklinks   = true, 
  colorlinks   = true, 
  urlcolor     = blue, 
  linkcolor    = blue, 
  citecolor    = blue 
}
\begin{document}

\maketitle

\begin{abstract}
The standard LLM training pipeline applies reinforcement learning (RL) only after pre-training and supervised fine-tuning (SFT). 
We question this status quo by training a LLM from scratch and applying RL, SFT, and SFT followed by RL directly to intermediate pre-training checkpoints. We find that RL is effective very early, and often matches the full SFT$\to$RL pipeline early as well. Through experiments on harder problems, we find that targeted pre-training data composition is a strong lever for RL effectiveness, even more so than model scale. Beyond reasoning accuracy, applying RL directly to base checkpoints expands the model's distribution; the sharpening effect reported in recent work arises only when RL follows SFT. The general capabilities of the model remain essentially unchanged by RL, while they degrade following SFT. Finally, we merge RL and SFT objectives by \textit{parallel averaging}, which outperforms across all other training methods discussed, across metrics, while preserving general capabilities. Together, these results suggest that LLM training might benefit from an expanded use of RL. 
\end{abstract}


\section{Introduction}
\label{sec:intro}
Until recently, the training recipe for Large Language Models (LLM) exclusively used the next-token prediction (NTP) objective via cross-entropy loss. However, with the advent of RL for language models~\citep{ouyang2022training, shao2024deepseekmath}, a newer advancement is the now-standard post-training phase which sequentially employs supervised finetuning (SFT) followed by RL. The NTP objective for pre-training and SFT is typically used over a static, external dataset, i.e., an \textit{off-policy} regime. Instead, for the RL objective, the model learns from its own \textit{on-policy} generations.


%
Under this standard training regime, RL training only occurs after a substantial amount of NTP training.
It is unclear whether this is fundamentally necessary for RL training
or simply a design choice~\citep{foster2025good}.
%
There has also been growing interest in changing this standard
and expanding the use of RL for pretraining~\citep{hatamizadeh2025rlp, li2025reinforcement, xing2025pretrainzero}.
In this work we attempt to answer a more fundamental question:
\vspace{4pt}

\centerline{\emph{When and how should an RL objective be used in LLM training?}}

While it has been widely observed that post-training
dramatically improves the reasoning of the model,
RL's influence on model capabilities has been
the subject of recent debate.
%
For example, a growing body of work argues that RL primarily
sharpens the model's existing output
distribution~\citep{yue2025limit-of-rlvr, wu2025invisible, karan2025reasoning, qin2025decomposingelementsproblemsolving}.
%
It is unclear whether these findings
are inherent to the RL objective or an artifact of
the standard training regime.
By studying various training objectives comprehensively
across stages of pre-training,
we shed light on a second fundamental question:
\vspace{4pt}

\centerline{\emph{What is the influence of RL on model capabilities?}}

To answer these questions, we perform a large-scale rigorous study of on-policy learning for LLM training.
We pretrain an LLM from scratch on a high-quality, reasoning-heavy corpus, saving checkpoints throughout the process.
For each base model checkpoint, we perform various different training runs:
 (\textbf{Direct RL}) RL on the base model checkpoint; (\textbf{SFT}) SFT using a single ground-truth demonstration per example; (\textbf{SFT-Gold}) SFT using multiple ground-truth demonstrations per example; and (\textbf{SFT$\rightarrow$RL}) RL on top of the SFT models, for both SFT and SFT-Gold, representing the standard LLM training pipeline.\looseness=-1

\begin{figure}[t]
    \centering
    \includegraphics[width=\linewidth]{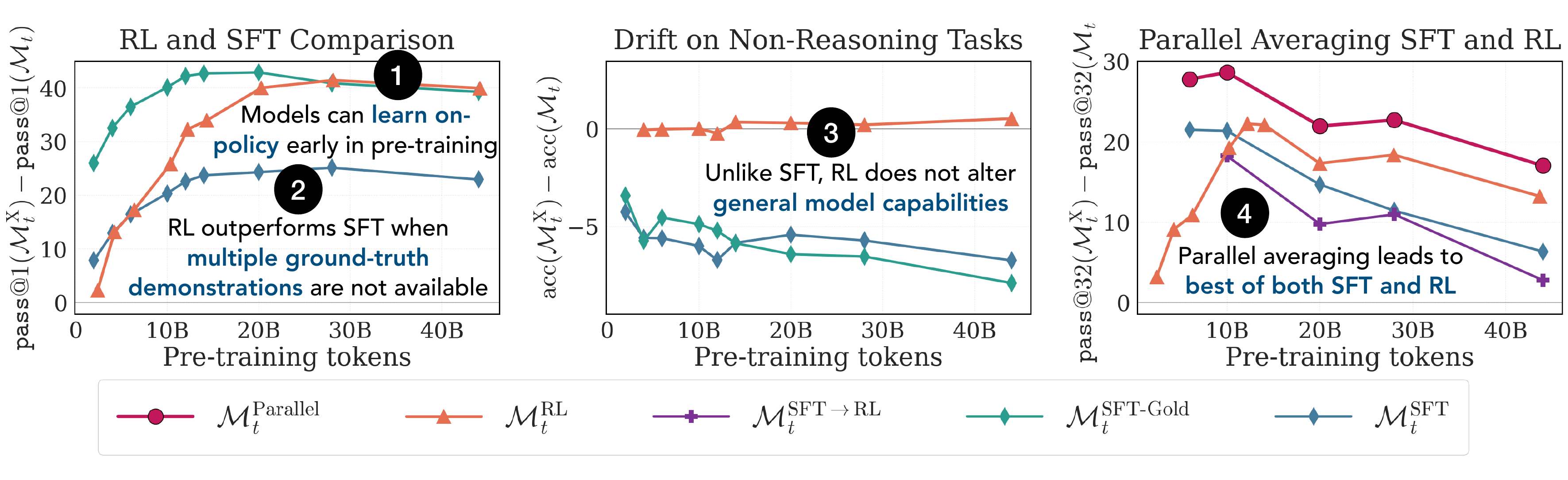}
    \vspace{-15pt}
    \caption{
      \textbf{Overview.}
      We compare several post-training recipes applied to intermediate pre-training checkpoints $\cM_t$: direct RL
  ($\mathcal{M}_t^{\text{RL}}$), SFT with one solution per question
  ($\mathcal{M}_t^{\text{SFT}}$), SFT with multiple solutions
  ($\mathcal{M}_t^{\text{SFT-Gold}}$), the standard pipeline of RL after SFT ($\mathcal{M}_t^{\text{SFT}\rightarrow\text{RL}}$), and parallel averaging of RL and SFT gradients ($\mathcal{M}_t^{\text{Parallel}}$).
      \figmark{1} RL improves both \texttt{pass@1} and \texttt{pass@32}
      on checkpoints trained for
      as low as 4B pretraining tokens (\S\ref{subsec:rlvr_gsm8k}).
      \figmark{2} RL is the more effective post-training objective when
      ground-truth demonstrations are scarce: $\mathcal{M}_t^{\text{RL}}$ substantially outperforms
      $\mathcal{M}_t^{\text{SFT}}$ on \texttt{pass@1} but matches  $\mathcal{M}_t^{\text{SFT-Gold}}$ (\S\ref{subsec:sft_multi}).\
      \figmark{3} SFT degrades general, non-reasoning benchmarks, whereas RL
      leaves these capabilities largely unchanged
      (\S\ref{subsec:general_capabilities}).
      \figmark{4} Parallel averaging of RL and SFT gradients combines their strengths: $\mathcal{M}_t^{\text{Parallel}}$ attains the strongest \texttt{pass@32} for every pre-training checkpoint (\S\ref{sec:interleaving}), consistently better than the standard SFT$\rightarrow$RL pipeline ($\mathcal{M}_t^{\text{SFT}\rightarrow\text{RL}}$).
    \vspace{-10pt}
    \label{fig:main_figure}}
\end{figure}

We present comprehensive findings that answer fundamental questions about RL for LLM training:

\textbf{When does RL work?} (\textbf{\S\ref{sec:main_results}}) We find that RL is effective surprisingly early in pretraining.
Training with direct RL on checkpoints that have seen as few as 4B tokens significantly improves performance on both GSM8K and MATH, with gains often comparable to the standard SFT$\rightarrow$RL pipeline (\S\ref{subsec:rlvr_gsm8k}).
Moreover, we find that RL is significantly more effective than SFT when we have limited
target demonstrations (SFT) (\S\ref{subsec:sft_multi}). The effectiveness of RL varies with task difficulty: RL gains are weaker on harder MATH-style problems.
In such cases, we find that adding targeted data to the pretraining corpus is effective and a better strategy than scaling model size (\S\ref{subsec:task_difficulty}).

\textbf{What does RL do?} (\textbf{\S\ref{sec:what}}) Contrary to recent claims that RL primarily sharpens the output distribution~\citep{yue2025limit-of-rlvr, wu2025invisible}, we find that RL applied directly to base checkpoints \emph{expands} the distribution: $\texttt{pass@1}$ \emph{and} $\texttt{pass@k}$ both improve substantially (\S\ref{subsec:rl_improves_passsk}). The sharpening effect we do reproduce arises only when RL is applied following SFT. This suggests that SFT, rather than RL itself, is what constrains exploration. Further, we find that SFT consistently degrades general (non-reasoning) capabilities, while RL leaves these capabilities unchanged (\S\ref{subsec:general_capabilities}).

\textbf{How should RL be used?} (\textbf{\S\ref{sec:interleaving}}) Finally, we investigate whether interleaving SFT and RL gradients within a single training step can capture the complementary strengths of both objectives. We propose a parallel-averaging update that combines updates from SFT and direct RL. We find that this simple objective yields better \texttt{pass@32} than all other recipes that use a single demonstration per problem, including SFT$\rightarrow$RL, indicating that using RL and NTP objectives simultaneously can be beneficial.

Overall, our findings make headway in understanding RL in contrast to other training objectives. 
Through our controlled experiments
across different stages of pre-training,
we find that a lot of assumptions about the RL objective
are artifacts of the current training regime.
Our results indicate that isolating the objectives
from the setting
reveal surprising aspects about RL as an objective for LLM training.
We focus our experiments on math reasoning capabilities to maintain a controlled training and evaluation environment.
We view our results as evidence that introducing RL earlier and more centrally in the LLM training pipeline is both feasible and, in several respects, preferable to the current standard.\looseness=-1

\section{Methodology and Experimental Design}
\label{sec:setup}

To answer foundational questions around the RL objective for LLM training, as stated above, beyond the standard training pipeline,
we first establish a controlled experimental environment.
Our setup centers on a custom-trained 1B model,
allowing for precise control over data exposure.
In this section, we detail pre-training checkpoints,
define three post-training training pipelines,
and describe data and evaluation.
While we explore RL as a general objective, we focus our implementation on Reinforcement Learning via Verifiable Rewards (RLVR) using the GRPO algorithm \citep{shao2024deepseekmath}.\looseness=-1

\subsection{Pre-training checkpoints}

\paragraph{Base model and data.}
We pre-train a $1$B parameter model based on OLMo2's~\citep{olmo20242} architecture and training infrastructure. We perform our pre-training from scratch using high-quality data based on a high-quality subset of OLMo2's pre-training mix, DOLMino \citep{olmo20242}\footnote{\hyperlink{https://huggingface.co/datasets/allenai/dolmino-mix-1124}{\url{allenai/dolmino-mix-1124}}}.
The DOLMino mix contains $50$B tokens, including general domains such as Wikipedia ($7$\%), high-quality web data ($60$\% from DCLM \citep{li2024datacomp} and FLAN \citep{weifinetuned}), high-quality math data ($20$\%), and other reasoning or code data such as StackExchange ($2$\%) and STEM papers $($5\%).\looseness=-1

\paragraph{Pre-training details.}
We pre-train our $1$B parameter model on $50$B tokens ($\sim2.5\times$ Chinchilla optimal tokens). 
We use AdamW~\citep{loshchilov2019} with a cosine learning rate decay and a peak learning rate of $4\times10^{-4}$. We train the model with a sequence length of $4096$ and batch size $512$. For experiments in Section~\ref{subsec:task_difficulty}, we perform two additional pre-training experiments. In the first, we keep the model architecture and size fixed, but add an additional $10$B tokens from the DOLMino-3 mixture~\citep{olmo2025olmo}\footnote{\hyperlink{https://huggingface.co/datasets/allenai/dolma3_dolmino_mix-100B-1125}{\url{allenai/dolma3\_dolmino\_mix-100B-1125}}} throughout training. In the second, we pre-train using the same $50$B tokens but scale the model size to $4$B parameters.
\looseness=-1

\subsection{Training Pipelines}
\label{sec:methods_and_baselines}
Let $\mathcal{M}_t$ denote the pre-training model checkpoint at step $t$, and $\mathcal{M}_T$ denote the final, fully-pre-trained model. We describe the three methods we compare below.

\begin{itemize} [leftmargin=*, itemsep=2pt, topsep=0pt, parsep=0pt]
    \item \textbf{Direct RL ($\cM_t^{\text{RL}}$)} We start with $\mathcal{M}_t$ and train with the RL objective. 
    \item \textbf{SFT only ($\mathcal{M}^{\text{SFT}}_t$):} We start with  $\mathcal{M}_t$ and perform SFT with ground-truth solutions. 
    \item \textbf{Standard pipeline ($\mathcal{M}^{\text{SFT} \rightarrow {\text{RL}}}_t$): } We train $\mathcal{M}^{\text{SFT}}_t$ with RL on the same set of questions.\looseness=-1
\end{itemize}

By comparing $\mathcal{M}^{\text{RL}}_t$ against $\mathcal{M}^{\text{SFT}}_t$, we isolate the training objective (RL vs. SFT) to determine if RL provides a superior training signal. By comparing $\mathcal{M}^{\text{RL}}_t$ with $\mathcal{M}^{\text{SFT} \rightarrow {\text{RL}}}_t$, we isolate if RL alone can provide a superior training signal than the standard pipeline.

In this work, we are interested in understanding, if given sufficient compute, how well each method performs. Therefore, we train all our RL and SFT runs until convergence, and we confirm the convergence of training in Appendix~\ref{appdx:gsm_rl_dynamics} and Appendix~\ref{appdx:sft_dynamics}.

\subsection{Data and Evaluation}

\paragraph{Training data.} We use OpenMathInstruct 
\citep{toshniwal2024openmathinstruct}, which consists of math 
problems paired with multiple ground-truth demonstrations per problem.
For SFT, by default,
we randomly pick a single solution per prompt for our training
(\textbf{SFT})
since that is a more realistic SFT setting
as obtaining multiple ground-truth reasoning traces
for each problem is typically infeasible.
However, we also consider training with all solutions (\textbf{SFT-Gold}) for completeness.
For RL, we only consider the final answer for each problem
and define a binary reward based on whether the model generation reaches the same final answer.

\paragraph{Difficulty splits.} OpenMathInstruct contains two 
categories of questions: a majority inspired by the MATH dataset 
\citep{hendrycksmath2021} (competition-level) and a minority 
inspired by GSM8K \citep{cobbe2021gsm8k} (grade-school level). 
We consider two experimental settings to probe different aspects of RL training: training with the full OpenMathInstruct and training with the GSM8K-inspired subset of OpenMathInstruct. On the GSM8K subset, the base pre-training checkpoints already achieve non-trivial performance. In contrast, the full MATH-heavy training set contains problems that remain challenging even for later pre-training checkpoints, allowing us to examine how far different pipelines can push the model's reasoning capabilities. 

\paragraph{Evaluation.} We evaluate on GSM8K and MATH 
respectively, reporting $\texttt{pass@k}$ 
\citep{Chen2021EvaluatingLL}, which estimates the probability of 
obtaining at least one correct response when $k$ responses are 
generated, for $k = 1, 8, 32$ and at temperature $T = 0.6$.

\section{RL is Effective Early in Pre-Training}
\label{sec:main_results}

\begin{figure}[t]
    \centering
    \includegraphics[width=1.0\linewidth]{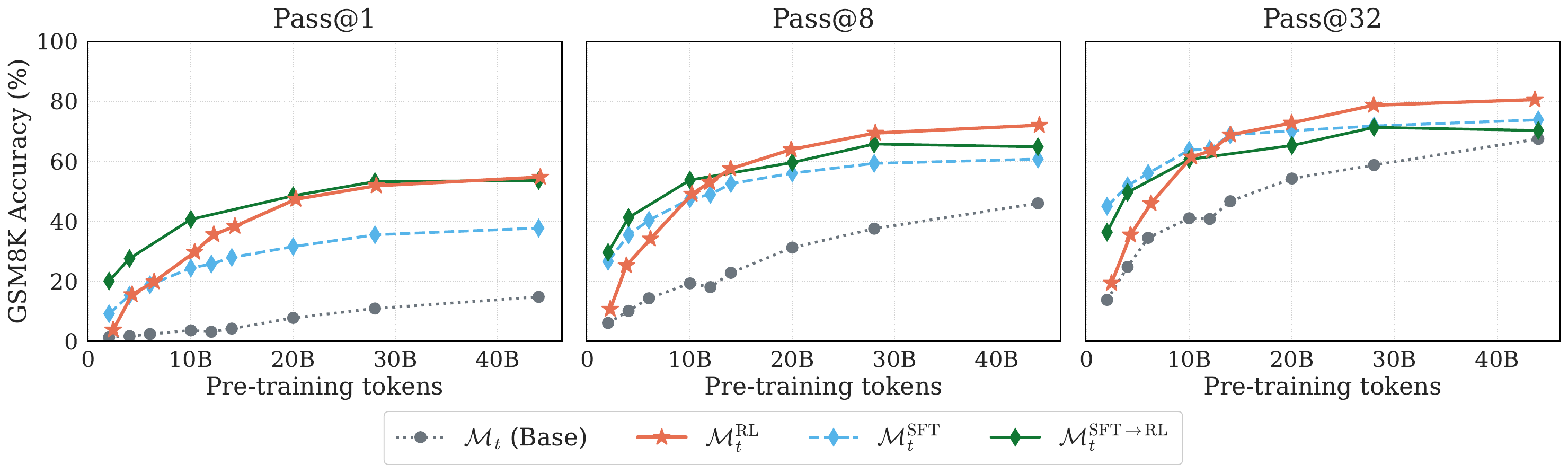}
    \caption{
    \textbf{RL is effective early in pre-training.}
    GSM8K $\texttt{pass@k}$ for $\cM_t$, $\cM_t^{\text{SFT}}$, $\cM_t^{\text{SFT} \to \text{RL}}$, and $\cM_t^{\text{RL}}$ across pre-training tokens $t$, with all SFT baselines trained on the SFT set (one ground-truth solution per problem). $\cM_t^{\text{RL}}$ improves over $\cM_t$ from as few as 4B tokens. By 10B tokens, $\cM_t^{\text{RL}}$ matches the standard $\cM_t^{\text{SFT}\to\text{RL}}$ pipeline, and outperforms $\cM_t^{\text{SFT}}$ alone.
    }
    \label{fig:omigsm_pre-train_to_rl_single}
\end{figure}

In this section, we study the effect of RL at different stages of pre-training and contrast with training objectives. On GSM8K, for our 1B parameter model, we find that RL is effective from as early as 4B pre-training tokens, and often matches the full SFT$\rightarrow$RL pipeline (\S\ref{subsec:rlvr_gsm8k}). RL is also the more effective objective when ground-truth demonstrations are scarce (\S\ref{subsec:sft_multi}). On harder problems, pre-training data composition is a stronger lever for RL effectiveness than model scale (\S\ref{subsec:task_difficulty}). Finally, we identify the the base model \texttt{pass@k} on the test set as a lightweight diagnostic for whether RL will succeed (\S\ref{sec:predicting_rl}).

\subsection{RLVR competes with the standard pipeline on GSM8K} 
\label{subsec:rlvr_gsm8k}

In \Cref{fig:omigsm_pre-train_to_rl_single}, we report the performance of $\mathcal{M}_t$, $\mathcal{M}_t^{\text{RL}}$, $\mathcal{M}_t^{\text{SFT}}$, and $\mathcal{M}_t^{\text{SFT} \rightarrow \text{RL}}$ at various pre-training steps $t$ on GSM8K, using the GSM8K subset of OpenMathInstruct for post-training. We evaluate base checkpoints $\mathcal{M}_t$ with 8-shot prompting, as they cannot reliably follow question-answering instructions\footnote{In Appendix~\ref{appdx:n_shot_eval}, we ablate the number of in-context examples and confirm that 8-shot yields the best performance for $\mathcal{M}_t$.}. All post-trained models use 0-shot evaluation, as RL includes a formatting reward and SFT data is formatted accordingly.

We observe that, as early as $t=4$B pre-training tokens, training with RL significantly improves the model's performance on GSM8K: for example, the \texttt{pass@1} accuracy increases from $\sim2\%$ to $\sim18\%$. 
Notably, the fact that this occurs at $t = 4$B tokens indicates \emph{improvement with RL prior to reaching the Chinchilla optimal number of tokens} \citep{hoffmann2022training}. In addition, we observe a significant increase in \texttt{pass@k} for $k=8,32$, which we discuss in detail in~\S\ref{subsec:rl_improves_passsk}.

\begin{figure}[t]
    \centering
    \includegraphics[width=1.0\linewidth]{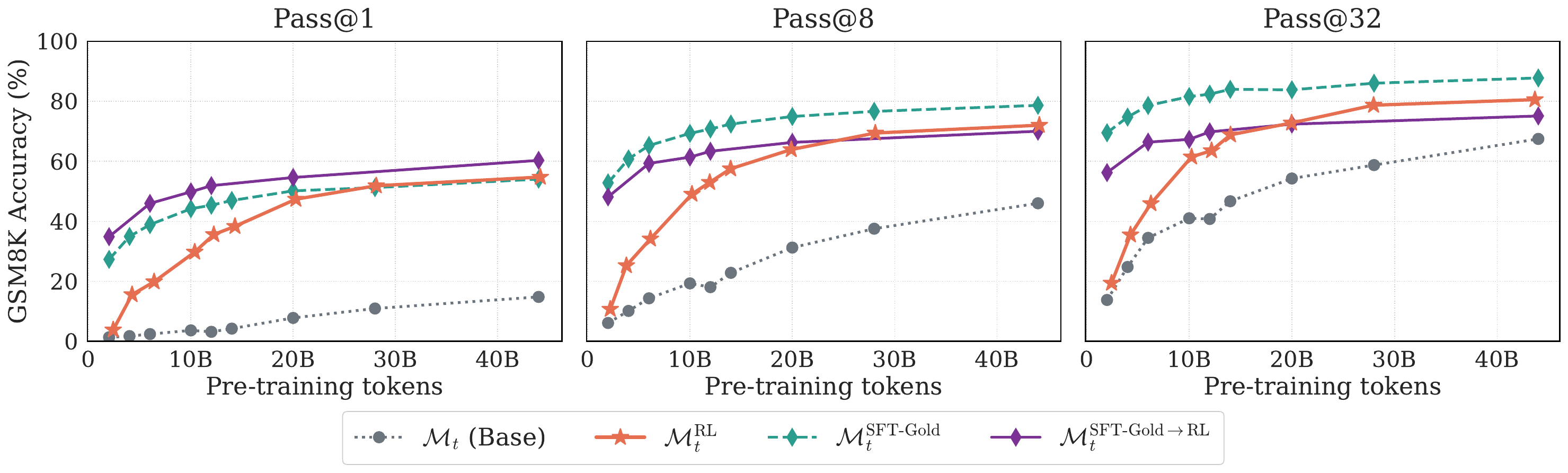}
    \caption{
    \textbf{Diverse SFT data shifts the balance toward SFT-Gold.}
    In contrast with \Cref{fig:omigsm_pre-train_to_rl_single}, SFT baselines are trained on SFT-Gold (all $\sim$23 ground-truth solutions per problem). With access to many ground-truth solutions, $\cM_t^{\text{SFT-Gold}}$ alone surpasses $\cM_t^{\text{RL}}$ on $\texttt{pass@8}$ and $\texttt{pass@32}$, while $\cM_t^{\text{SFT-Gold}\to\text{RL}}$ remains best on $\texttt{pass@1}$. $\cM_t^{\text{SFT-Gold}}$'s advantage requires multiple high-quality solutions per problem, which is rarely realistic in practice.
    }
    \label{fig:omigsm_pre-train_to_rl_multi}
\end{figure}

In \Cref{fig:omigsm_pre-train_to_rl_multi}, after $t=10$B tokens, $\mathcal{M}_t^{\text{RL}}$ \textit{outperforms} $\mathcal{M}_t^{\text{SFT}}$ on $\texttt{pass@1}$ and performs on-par with $\mathcal{M}_t^{\text{SFT} \rightarrow \text{RL}}$. For $\texttt{pass@8,32}$, $\mathcal{M}_t^{\text{RL}}$ performs on-par with both $\mathcal{M}_t^{\text{SFT}}$ and $\mathcal{M}_t^{\text{SFT} \rightarrow \text{RL}}$. This result is significant because $\mathcal{M}_t^{\text{RL}}$ never observes ground-truth reasoning traces; unlike the SFT baselines, it develops reasoning capabilities entirely from self-generated traces and feedback, demonstrating that RL can match supervised learning without training on ground-truth reasoning traces.

For some early $\mathcal{M}_t$ model checkpoints between $t=4$B and $t=10$B pre-training tokens, we observe that for brittleness across seeds: RL performance on some seeds fails to improve. It is likely that the model sometimes falls into a distinct failure mode for early pre-training checkpoints (Appendix~\ref{appdx:seed_dependency}).
Above, we report the RL runs with non-trivial performance.
\subsection{RL outperforms when SFT data is scarce} 
\label{subsec:sft_multi}

OpenMathInstruct contains an average of 23 ground-truth completions per problem. Our main results use the one randomly chosen completion per problem, which we consider the more realistic SFT setting. We also consider a setting that uses the full OpenMathInstruct dataset for SFT training (i.e., multiple ground-truth completions per problem). We refer to this setting as SFT-Gold since obtaining multiple high-quality solutions per problem typically requires expensive human supervision or generation from frontier models, making it an ideal setting which is impractical for many domains. \Cref{fig:omigsm_pre-train_to_rl_single} shows that $\mathcal{M}_t^{\text{RL}}$ outperforms $\mathcal{M}_t^{\text{SFT}}$ on \texttt{pass@1} and is competitive on \texttt{pass@8,32}. With SFT-Gold, the story changes (\Cref{fig:omigsm_pre-train_to_rl_single}): SFT-Gold$\rightarrow$RL performs best on \texttt{pass@1}, but SFT-Gold alone surpasses both RL and SFT$\rightarrow$RL on \texttt{pass@8,32}. This suggests that access to diverse ground-truth reasoning traces can provide coverage benefits that on-policy exploration does not.

\subsection{Targeted pre-training data is more essential than model size for RL} 
\label{subsec:task_difficulty}

In \Cref{fig:omi_pretrain_to_rl_1b}, we train on the MATH-like subset of OpenMathInstruct and evaluate \texttt{pass@k} accuracy on MATH. Unlike the GSM8K setting in \Cref{fig:omigsm_pre-train_to_rl_single}, directly applying RL to pre-training checkpoints is less effective with respect to SFT with SFT and SFT-Gold on this harder benchmark.  We hypothesize that the \textit{efficacy} of RL training from pre-training checkpoints has limitations, potentially related to the difficulty of the task at hand.  Therefore, we study two natural interventions: scaling $N$, the model size, and scaling $D$, the amount of pre-training data, especially task-relevant math data.


\begin{figure}[t]
    \centering
    \includegraphics[width=\linewidth]{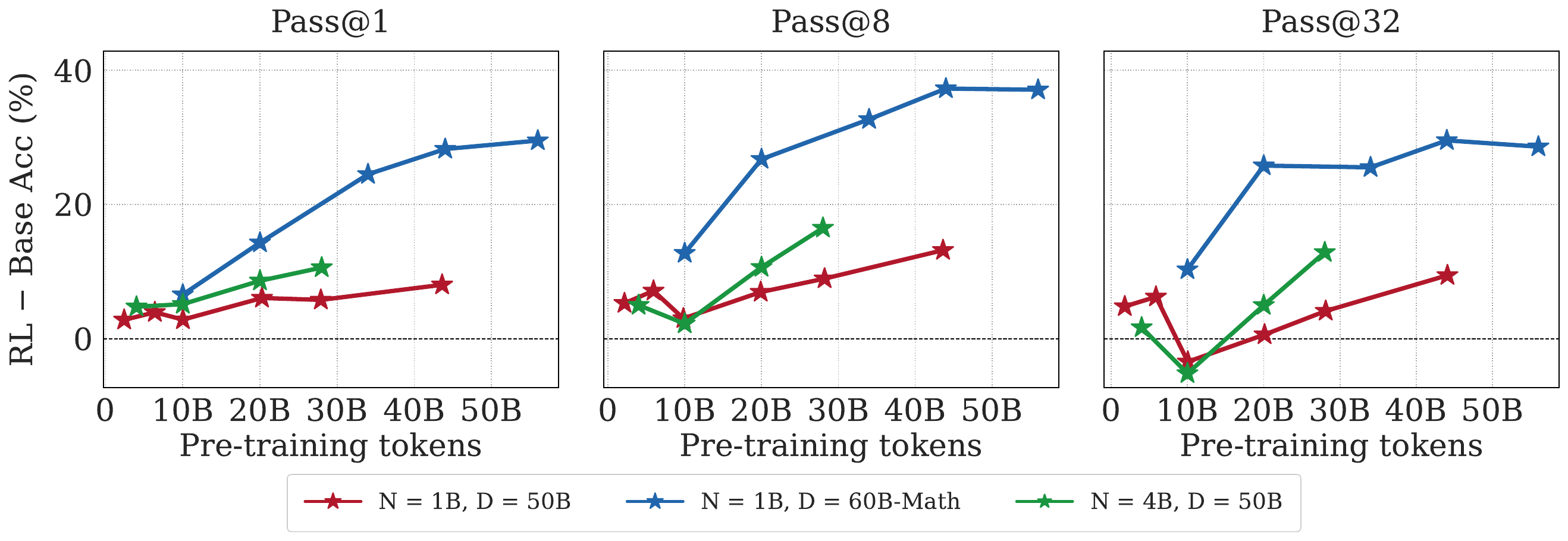}
    \caption{\textbf{Targeted pre-training data beats model scale for RL.} Improvement on MATH from RL over the base model, across pre-training configurations: (i) \textit{1B-50B}, original pre-trained model; (ii) \textit{Scaling $D$ (1B-60B)}, 1B model pre-trained from scratch with an additional 10B math-heavy tokens mixed in; (iii) \textit{Scaling $N$ (4B-50B)}, 4B model trained on same 50B-token mix as original 1B model. Adding task-relevant pre-training data (Scaling $D$) yields substantially larger RL gains on MATH.}
    \label{fig:math_rl_comparison}
\end{figure}



To test the effect of scaling $N$, we pre-train a 4B model from scratch using the same 50B-token mix and training recipe as the original 1B model. As expected, the 4B checkpoints achieve higher base MATH accuracy, and direct RL on these checkpoints also yields higher absolute performance than RL on the 1B checkpoints at matched pre-training steps. However, when measuring the gain from RL relative to each checkpoint's own base performance, the 4B model does \textit{not} obtain larger improvements. As shown in \Cref{fig:math_rl_comparison}, scaling model size improves the base model, but does not substantially improve the effectiveness of RL itself.


We then test the effect of scaling $D$ while keeping the model size fixed at 1B. We pre-train from scratch with an additional 10B math- and reasoning-heavy tokens from the Dolma 3 Dolmino Mix \citep{olmo2025olmo}, described in Appendix~\ref{app:10b_new_tokens}. In this setting, direct RL on the resulting checkpoints matches the SFT baseline on MATH (\Cref{fig:omi_pretrain_to_rl_1b_60b}) and recovers the qualitative behavior observed on GSM8K (\Cref{fig:omigsm_pre-train_to_rl_single}). Moreover, \Cref{fig:math_rl_comparison} shows that the RL gain over the base model is substantially larger than in either the original 1B setting or the 4B scaling-$N$ setting.

Overall, targeted pre-training data is the more effective intervention: adding math-specific data during pre-training substantially improves the gains achievable by direct RL, whereas increasing model size primarily improves the base checkpoint. In Appendix~\ref{sec:rollouts}, we further study scaling $G$, the number of RL rollouts, and find that increasing $G$ does \textit{not} change the final outcome of direct RL.


\subsection{Base model performance is predictive of RL effectiveness}
\label{sec:predicting_rl}

\begin{figure}[t]
    \centering
    \includegraphics[width=\linewidth]{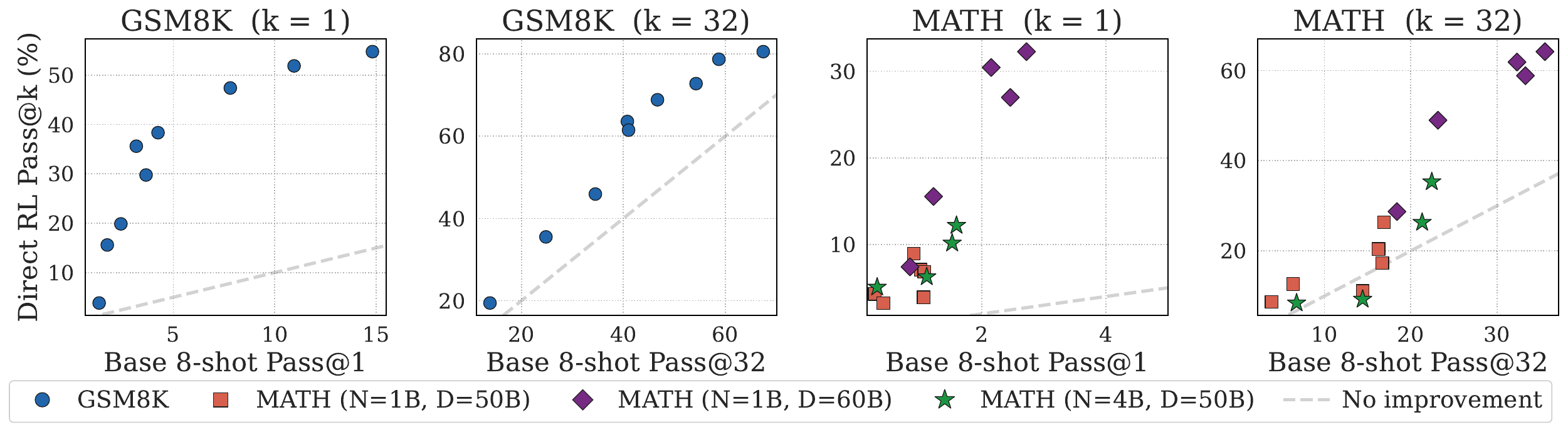}
    \caption{
    \textbf{Base $\texttt{pass@k}$ on training data predicts RL effectiveness.} Base model 8-shot $\texttt{pass@k}$ on the test set ($x$-axis) vs. after RL ($y$-axis), for GSM8K(\textit{left}) and MATH (\textit{right}). \texttt{pass@k} accuracy on the test set might serve as a lightweight metric for whether RL training will yield downstream gains.
    }
    \label{fig:base_to_rl_improvement}
\end{figure}

Given that we train with RL on early pre-training checkpoints, a natural question is, \textit{how can we predict if RL training will be effective? }
%
In Figure~\ref{fig:base_to_rl_improvement}, we compare the base model's \texttt{pass@k} accuracy on the test set with that of the model after RL. For MATH, we also report the comparison for the two additional pre-training regimes discussed in \S\ref{subsec:task_difficulty}. We observe a generally monotonically increasing relationship in which increasing \texttt{pass@k} for the base model corresponds to increased\texttt{pass@k} for the model after RL. In practice, this suggests that a model's \texttt{pass@k} accuracy on the test set might serve as a lightweight metric for whether RL training will yield downstream gains.

\section{The Effects of RL Beyond Downstream Accuracy}
\label{sec:what}

\begin{figure}[t]
    \centering
    \includegraphics[width=0.9\linewidth]{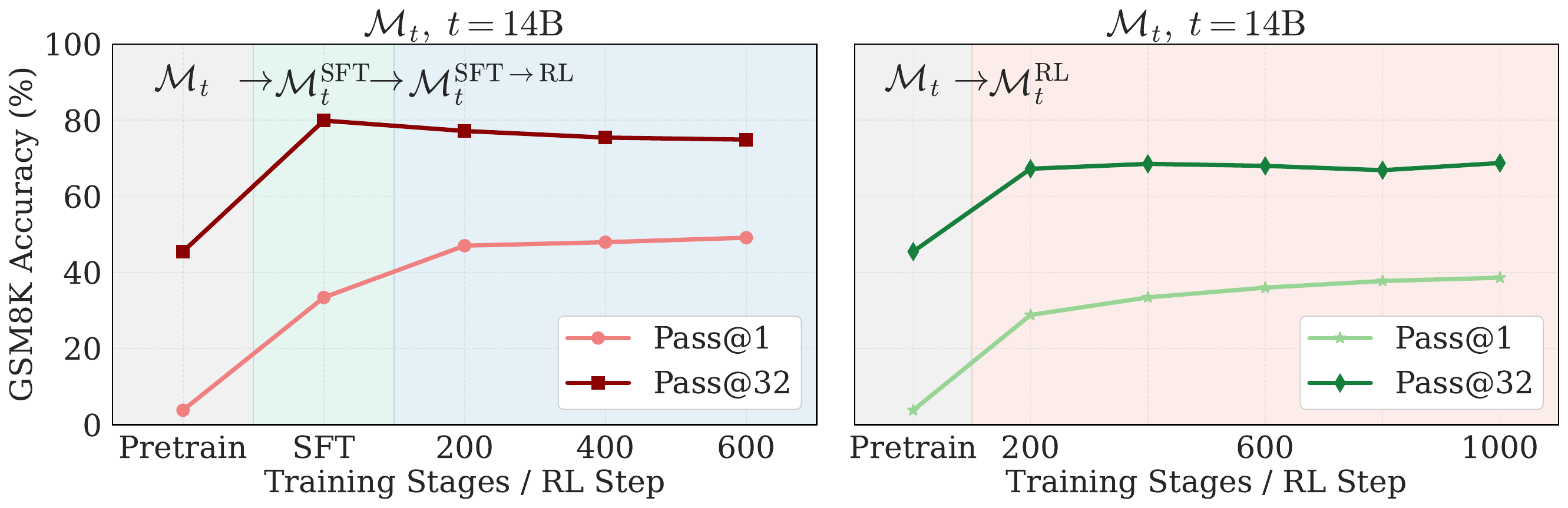}
    \caption{\textbf{Direct RL expands while SFT$\to$RL sharpens.} GSM8K $\texttt{pass@1}$ and $\texttt{pass@32}$ tracked across training stages on the same pretraining checkpoint $\cM_t$. \textit{Left}: under the standard $\cM_t \to \cM_t^{\text{SFT}} \to \cM_t^{\text{SFT}\to\text{RL}}$ pipeline, $\texttt{pass@1}$ continues to improve during RL but $\texttt{pass@32}$ \emph{decreases}, reproducing the sharpening effect reported in prior work. \textit{Right}: applying RL directly to $\cM_t$ improves both $\texttt{pass@1}$ and $\texttt{pass@32}$, expanding the model's distribution rather than merely sharpening it.
    }
    \label{fig:rl_sft_dynamics}
\end{figure}

\begin{figure}[t]
    \centering
    \includegraphics[width=0.85\linewidth]{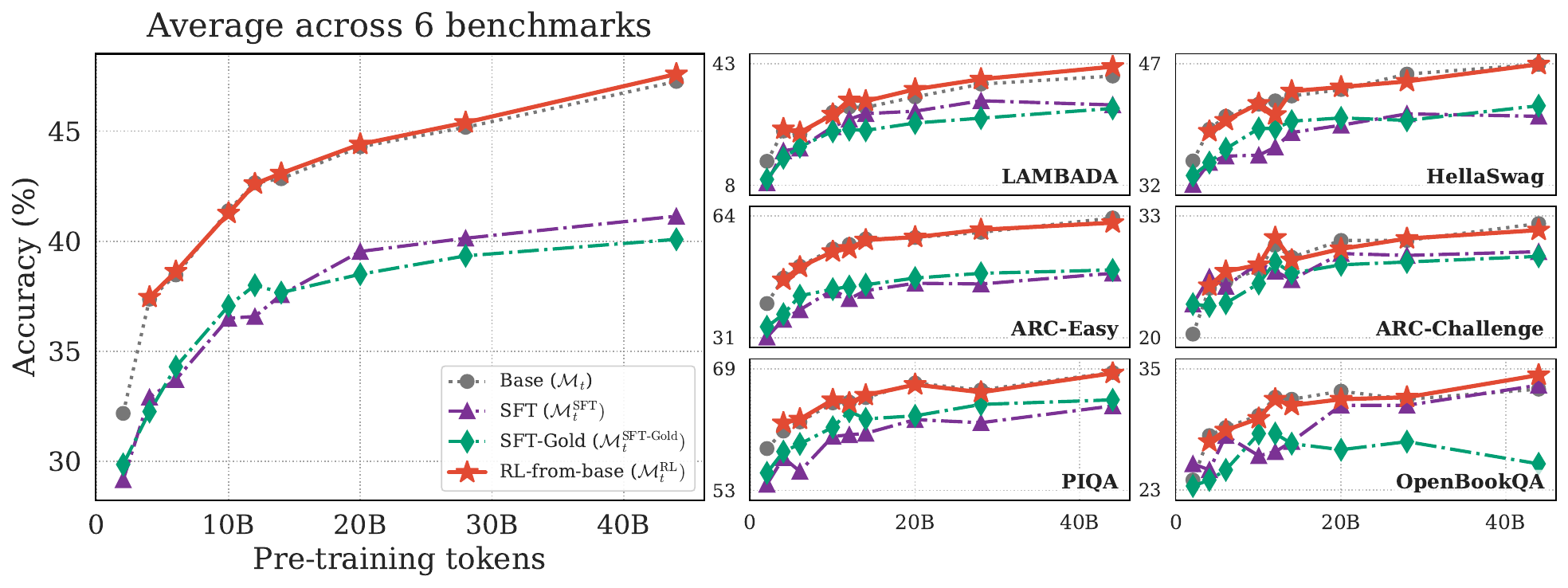}
    \caption{
    \textbf{RL preserves general capabilities while SFT degrades them.} Performance on six general-purpose (non-math) benchmarks for the base model $\cM_t$ and three post-trained variants: $\cM_t^{\text{RL}}$, $\cM_t^{\text{SFT}}$, and $\cM_t^{\text{SFT-Gold}}$. Both SFT and SFT-Gold consistently degrade performance by $4$--$8$\,pp on average across the benchmarks, while RL leaves these capabilities essentially unchanged.
    }
    \label{fig:general_regression}
\end{figure}

In this section, we examine the effects of RL on the trained model beyond the downstream accuracy. First, we evaluate whether RL unlocks new reasoning capabilities or merely \emph{sharpens} the base model's existing output distribution (\S\ref{subsec:rl_improves_passsk}). Then, we study whether RL alters general capabilities inherited from pretraining (\S\ref{subsec:general_capabilities}). We find that the answer to both depends on the training pipeline. 

\subsection{Early stage RL can expand the model's distribution}
\label{subsec:rl_improves_passsk}

We refer to  sharpening~\citep{wu2025invisible, yue2025limit-of-rlvr} as the phenomenon in which training improves $\texttt{pass@1}$ accuracy but has little or even negative effect on $\texttt{pass@k}$ accuracy for larger $k$. In contrast, we define expansion as the setting in which RL increases $\texttt{pass@k}$ performance across for large $k$.\looseness=-1

Many recent works have claimed that RLVR largely \textit{sharpens} the distribution without bringing the model any ``new" reasoning capabilities \citep{yue2025limit-of-rlvr, Cheng2026isocompute}. These works point to evidence that during RL, $\texttt{pass@k}$ does not improve for sufficiently large $k$. Interestingly, in our experiments we observe two opposing outcomes depending on the training pipeline. First, when we apply the standard pipeline on pretraining checkpoints (i.e., $\mathcal{M}_t \rightarrow \mathcal{M}_t^{\text{SFT}} \rightarrow \mathcal{M}_t^{\text{SFT} \rightarrow \text{RL}}$), we observe the sharpening effect. In \Cref{fig:rl_sft_dynamics} (\textit{left}), we show one such example. We see that $\texttt{pass@1}$ continues to improve from $\cM_t$ to $\cM_t^\text{SFT}$, and then to $\cM_t^\text{SFT}\rightarrow\text{RL}$. On the other hand, the SFT stage yields a significant gain in $\texttt{pass@32}$, but the subsequent RL stage slightly degrades the performance.

We hypothesize that sharpening occurs because, during SFT, the model has already seen ground-truth solutions on the same set of questions, thus RL primarily refines these existing capabilities rather than discovering new reasoning paths. In contrast, by directly training on the RL objective from the same pretraining checkpoint \Cref{fig:rl_sft_dynamics} (\textit{right}), RL training improves both $\texttt{pass@1}$ and $\texttt{pass@32}$ performance, expanding the base model's distribution. Without prior exposure to ground-truth solutions, the model explores and discovers new reasoning paths through on-policy learning.

\subsection{RL does not affect general model capabilities}
\label{subsec:general_capabilities}

%
A natural concern with applying RL to intermediate pretraining checkpoints is whether it degrades capabilities outside the training domain. To assess this, we evaluate on several general-purpose benchmarks and report results in Figure~\ref{fig:general_regression}. We report benchmark results for the base model, the base model after training directly with RL, and training with SFT. For SFT, we report accuracy both for training with SFT and SFT-Gold. 
Interestingly, we find that RL from the base model has the least effect on general model capability, while SFT training typically degrades the general model capability regardless of using one or many completions per prompt.

\section{Parallel RL and SFT}
\label{sec:interleaving}

\begin{figure}[t]
\centering
\begin{minipage}[c]{0.46\textwidth}
\captionsetup{
    type=algorithm,
    font=small,
    labelfont=bf,
    labelsep=space,
    justification=raggedright,
    singlelinecheck=false
}

\hrule height 1.2pt
\vspace{0.35em}
\caption{Parallel averaging update}
\label{alg:parallel_avg}
\vspace{-0.6em}
\hrule

\small
\begin{algorithmic}[1]
\STATE \textbf{Input:} parameters $\theta$; optimizer states $s_{\text{RL}}, s_{\text{SFT}}$;
batches $\mathcal{B}_{\text{RL}}, \mathcal{B}_{\text{SFT}}$; learning rates $\eta_{\text{RL}}, \eta_{\text{SFT}}$
\ALGComment{Snapshot current parameters}
\STATE $\bar{\theta} \leftarrow \theta$
\ALGComment{Compute both objectives at the same snapshot}
\STATE $g_{\text{RL}} \leftarrow
\nabla_{\theta}\mathcal{L}_{\text{RL}}(\theta;\mathcal{B}_{\text{RL}})
\big|_{\theta=\bar{\theta}}$
\STATE $g_{\text{SFT}} \leftarrow
\nabla_{\theta}\mathcal{L}_{\text{SFT}}(\theta;\mathcal{B}_{\text{SFT}})
\big|_{\theta=\bar{\theta}}$

\ALGComment{Compute optimizer updates from the snapshot}
\STATE $(\Delta_{\text{RL}}, s_{\text{RL}}) \leftarrow
\mathrm{OptUpdate}(g_{\text{RL}}, s_{\text{RL}}, \eta_{\text{RL}})$
\STATE $(\Delta_{\text{SFT}}, s_{\text{SFT}}) \leftarrow
\mathrm{OptUpdate}(g_{\text{SFT}}, s_{\text{SFT}}, \eta_{\text{SFT}})$

\ALGComment{Average the two gradient updates}
\STATE $\theta \leftarrow \bar{\theta}
+ \tfrac{1}{2}\left(\Delta_{\text{RL}}+\Delta_{\text{SFT}}\right)$

\STATE \textbf{return} $\theta$
\end{algorithmic}

\hrule
\end{minipage}%
\hfill
\begin{minipage}[c]{0.52\textwidth}
\centering
\includegraphics[width=\linewidth]{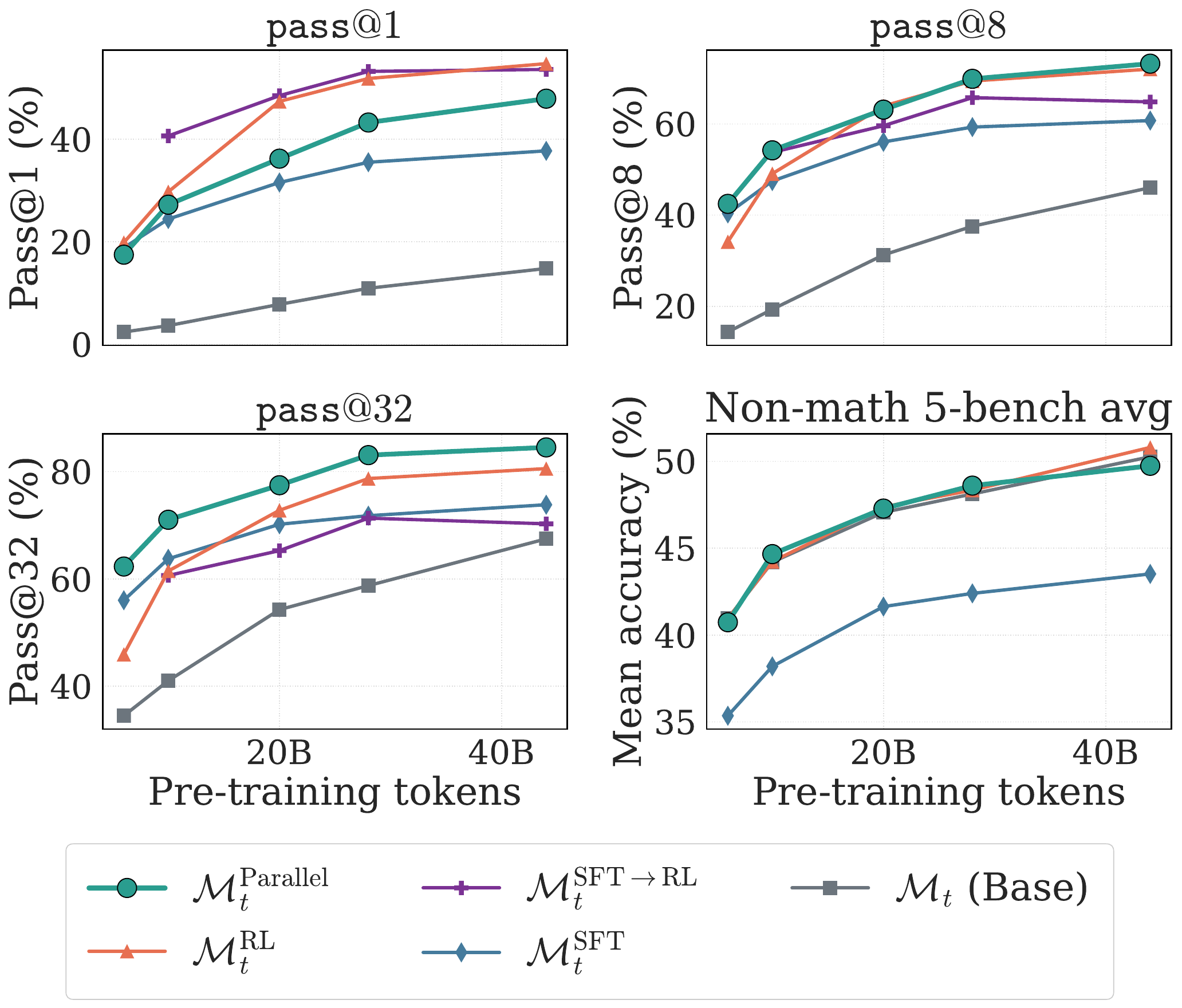}
\end{minipage}
\caption{\textbf{Parallel averaging combines the strengths of RL and SFT across pre-training.} \textit{(Left)} The parallel-averaging update (\Cref{alg:parallel_avg}): at each step we take a single optimizer update from each of an RL gradient and an SFT gradient (each with its own optimizer state) and use their average to update the model weights. \textit{(Right)} Parallel-averaging ($\mathcal{M}_t^{\text{Parallel}}$) achieves the strongest $\texttt{pass@32}$ across pre-training checkpoint surpassing the standard pipeline ($\mathcal{M}_t^{\text{SFT}\rightarrow\text{RL}}$). Unlike SFT-based regimes, $\mathcal{M}_t^{\text{Parallel}}$ does not regress on non-math benchmarks and retains base model performance.
\label{fig:parallel-avg-main}}
\end{figure}

The previous sections expose complementary strengths of the RL and SFT objectives applied directly to pretraining checkpoints. Direct RL ($\mathcal{M}_t^{\text{RL}}$) can develop new reasoning capabilities, expand the model's $\texttt{pass@k}$ distribution, and leave general (non-math) capabilities intact (\S\ref{sec:what}). This expansion, however, is only reliable when the underlying pretraining mix and model size yield enough latent capability to bootstrap from the base model (\S\ref{subsec:task_difficulty}). SFT instead provides reliable supervision from ground-truth reasoning traces. 
However, the efficacy of SFT relies on the diversity of the SFT data, and might have a negative impact on general capabilities. Having studied RL and SFT in isolation, we next consider whether a combined objective might enjoy the benefits of both. 

\paragraph{Method.}
We propose a simple algorithm that combines both training objectives (Algorithm~\ref{alg:parallel_avg}). At each training step, starting from the same parameter snapshot $\theta$, we run one batch through an RL optimizer, and in parallel, run a separate batch of SFT data through an SFT optimizer. We obtain gradients from the two optimizers and average the gradients to update $\theta$. Critically, the two optimizers maintain independent first- and second-moment estimates, so the adaptive step sizes and preconditioning do not interfere.
We refer to the resulting model as $\mathcal{M}_t^{\text{Parallel}}$. 


\paragraph{Findings.}
We report results in \Cref{fig:parallel-avg-main} (see \Cref{fig:paralle_avg_full} for the per-checkpoint training trajectories). Across every pre-training checkpoint we evaluated, parallel averaging attains the strongest $\texttt{pass@32}$ among recipes that use a single demonstration per problem, surpassing direct RL, SFT, and the standard pipeline. It also preserves the base model's general (non-math) capabilities on par with direct RL, whereas every SFT-based recipe regresses on this axis by 5--8 percentage points. However, we also observe that this strong $\texttt{pass@k}$ improvements come with a trade-off of a lower $\texttt{pass@1}$ relative to the direct RL and SFT baselines.

Overall, we read these results as evidence that the RL and SFT signals are complementary rather than merely additive: the SFT loss supplies supervisory structure on reasoning paths that on-policy rollouts may rarely sample, while the concurrent RL signal anchors the model to its base distribution and avoids the general-capability regression typically seen after a dedicated SFT stage. Our recipe uses equal-weight averaging with no scheduling, leaving room for more deliberate combinations of RL and next-token-prediction objectives which we view as a productive direction for future work.
%

\section{Prior Work}
\label{sec:related}
\paragraph{Reinforcement Learning for LLM Reasoning}
RL has become a standard post-training stage for LLMs~\citep{ouyang2022training, dai2023safe, jaech2024openai}, using modern policy-gradient methods~\citep{rafailov2023direct, shao2024deepseekmath, yu2025dapo, khatri2025artscalingreinforcementlearning} with verifiable rewards~\citep{guo2025deepseek, zheng2023secrets}. Whether gains via these methods reflect new capabilities or merely a sharpeneing remains contested~\citep{yue2025limit-of-rlvr, wu2025invisible, karan2025reasoning, Cheng2026isocompute, chu2025sft}, as does whether RL erodes abilities inherited from pretraining~\citep{shenfeld2025rl}. Our findings suggest a nuanced view on these questions.\looseness=-1

\paragraph{Integrating RL into Pretraining}
A recent line of work brings RL into pretraining itself, either by scoring next-sentence reasoning against the training corpus~\citep{li2025reinforcement} or by inserting chain-of-thought rollouts before each next-token prediction~\citep{hatamizadeh2025rlp, dong2025reinforcement, xing2025pretrainzero}. These methods modify the pretraining objective. Our work instead keeps the standard NTP and RL objectives unchanged. We view our results as a precursor: before adding RL \emph{into} pretraining, it is worth knowing how early in pretraining RL on top of NTP already pays off.

\paragraph{Interleaving SFT and RL}
A growing body of work performs mixed-policy training. Approaches include importance-weighted off-policy expert traces~\citep{yan2025learning}, alternating SFT and RL passes targeted at unsolved problems~\citep{dong2025rl}, joint losses with adaptive weighting~\citep{fu2025srft, lv2025towards, zhang2026onpolicyrlmeetsoffpolicy}, and hybrid trajectories that blend expert prefixes with on-policy continuations~\citep{huang2025blending}. \citet{limozin2026sft} caution that several of these methods were compared against deflated SFT baselines, and that a correctly implemented SFT$\to$RL pipeline can match or exceed them. In our work, we evaluate a simple alternate approach that maintains independent Adam moments for SFT and RL and average their proposed updates after each step.\looseness=-1

\paragraph{Prerequisites for Post-Training}
A small but growing literature studies what level of pretraining is required before post-training becomes effective. \citet{chen2025coverage, foster2025good} argue that the base model must reach a minimum capability for RL to yield gains. \citet{zhang2025interplay} relates this threshold to the base model's basic skills and to the difficulty of the RL data; see also \citet{guo2025deepseek, zhou2023lima}. Our work tests this premise empirically by tracing RL effectiveness across pretraining tokens, and finds that the threshold is far lower than commonly assumed: RL is effective on GSM8K from as few as 4B tokens, well below the Chinchilla-optimal point.
\section{Discussion \& Future Directions}

In this work, we provide a comprehensive and nuanced picture of the RL objective for LLM training beyond how it is used in the current standard pipeline.
We find that RL can be effective starting early in pre-training, well before the Chinchilla-optimal regime, and often matches the full SFT$\to$RL pipeline on GSM8K tokens despite never seeing a ground-truth reasoning trace.
Further, the dominant lever for whether early RL succeeds is pre-training data composition, not model scale.
Perhaps most strikingly, the two effects most commonly attributed to RL, distribution sharpening and regression on general capabilities, are largely artifacts of a preceding SFT stage rather than of the RL objective itself: applied directly to base checkpoints, RL instead expands the $\texttt{pass@k}$ distribution and leaves non-math capabilities essentially intact, unlike SFT that consistently degrades them significantly.

Our results open several exciting research directions. The most consequential is rethinking how data and objectives are coordinated end-to-end: if RL is effective well inside pre-training, and the pre-training mix controls its ceiling, the practical question becomes what pre-training recipes could look like once RL is treated as a first-class training objective rather than a final post-training step.
Our parallel-averaging experiment (\S\ref{sec:interleaving}) is an early data point towards that end. Work can be done towards more careful designs, with adaptive weighting, scheduling, or importance sampling on top of independent optimizer states.
The finding that base $\texttt{pass@k}$ already predicts RL effectiveness (\S\ref{sec:predicting_rl}) further motivates adaptive rollout strategies that concentrate compute on prompts where the base model has non-trivial coverage but has not yet converged. Our experiments are at 1B and 4B parameters and 50--60B tokens; whether the same picture holds at frontier scale is essential future work.\looseness=-1

\paragraph{Limitations} We discuss a few limitations in our work. First, while our pre-training mix is designed to be reflective of general pre-training settings, it is more math-heavy than typical web-scale corpora. Second, we focus on standard GRPO as a representative RLVR objective and do not study the growing family of variants (e.g., those explicitly targeting entropy preservation or $\texttt{pass@k}$ expansion), which may interact differently with the pre-training stage.

\clearpage
\bibliographystyle{ICLR/iclr2026_conference}
\bibliography{references}

\clearpage
\appendix

\makeatletter
\def\addcontentsline#1#2#3{%
  \addtocontents{#1}{\protect\contentsline{#2}{#3}{\thepage}{page.\thepage}}%
}

\let\old@sect\@sect
\def\@sect#1#2#3#4#5#6[#7]#8{%
  \phantomsection 
  \old@sect{#1}{#2}{#3}{#4}{#5}{#6}[#7]{#8}%
}

\let\old@ssect\@ssect
\def\@ssect#1#2#3#4#5{%
  \phantomsection
  \old@ssect{#1}{#2}{#3}{#4}{#5}%
}
\makeatother

\setcounter{tocdepth}{2}

\begingroup
\setlength{\parskip}{0pt}
\setlength{\baselineskip}{0pt}
\renewcommand{\baselinestretch}{1.0}
\tableofcontents
\endgroup
\newpage

\section{Experiment Details}
\label{app:experiment_details}

In this section, we provide the details necessary to replicate our experiments. For pretraining, we use the Olmo pretraining library, and for RL/ SFT we use the VeRL library. 

\subsection{Resources}
For our experiments, we use a combination of NVIDIA A100 GPUs and NVIDIA H100 GPUs. Pretraining takes several days, GRPO training takes several days, and SFT takes a few hours. 

\subsection{Hyperparameters}
In the following tables, we report hyperparameter choices for GRPO, SFT, and pretraining.

\begin{table}[h]
\centering
\caption{GRPO training hyperparameters (OLMo2-1B on GSM8K subset).}
\label{tab:hyperparams}
\begin{tabular}{lll}
\toprule
\textbf{Category} & \textbf{Hyperparameter} & \textbf{Value} \\
\midrule
\multirow{4}{*}{Data}
  & Train batch size       & 512 \\
  & Max prompt length      & 1024 \\
  & Max response length    & 2048 \\
  & Rollouts per prompt ($n$) & 32 \\
\midrule
\multirow{4}{*}{Optimization}
  & Learning rate          & $1 \times 10^{-6}$ \\
  & Optimizer                    & AdamW \\
  & $(\beta_1, \beta_2)$         & $(0.9,\ 0.999)$ \\
  & Weight decay                 & 0.01 \\
  & Gradient clip                & 1.0 \\
  & Mini-batch size        & 128 \\
  & KL loss coefficient    & $1 \times 10^{-3}$ \\
  & KL loss type           & low-variance KL \\
\midrule
\multirow{2}{*}{Reward}
  & Advantage estimator    & GRPO \\
  & Format score (partial) & 0.1 \\
\midrule
\multirow{2}{*}{Infrastructure}
  & Total epochs           & 10 \\
  & GPU memory utilization & 0.6 \\
\bottomrule
\end{tabular}
\end{table}

\begin{table}[htbp]
\centering
\caption{GRPO training hyperparameters (OLMo2-1B on OpenMathInstruct-2).}
\label{tab:hyperparams_omi}
\begin{tabular}{lll}
\toprule
\textbf{Category} & \textbf{Hyperparameter} & \textbf{Value} \\
\midrule
\multirow{4}{*}{Data}
  & Train batch size             & 512 \\
  & Max prompt length            & 1024 \\
  & Max response length          & 2048 \\
  & Rollouts per prompt ($n$)    & 32 \\
\cmidrule{1-3}
\multirow{4}{*}{Optimization}
  & Learning rate                & $1 \times 10^{-6}$ \\
  & Optimizer                    & AdamW \\
  & $(\beta_1, \beta_2)$         & $(0.9,\ 0.999)$ \\
  & Weight decay                 & 0.01 \\
  & Gradient clip                & 1.0 \\
  & KL loss coefficient          & $1 \times 10^{-3}$ \\
  & KL loss type                 & low-variance KL \\
\cmidrule{1-3}
\multirow{2}{*}{Reward}
  & Advantage estimator          & GRPO \\
  & Format score (partial)       & 0.1 \\
\cmidrule{1-3}
\multirow{2}{*}{Infrastructure}
  & Total epochs                 & 10 \\
  & GPU memory utilization       & 0.8 \\
\bottomrule
\end{tabular}
\end{table}

\begin{table}[htbp]
\centering
\caption{SFT training hyperparameters (OLMo2-1B on OpenMathInstruct-2).}
\label{tab:hyperparams_sft}
\begin{tabular}{lll}
\toprule
\textbf{Category} & \textbf{Hyperparameter} & \textbf{Value} \\
\midrule
\multirow{4}{*}{Data}
  & Train batch size             & 512 \\
  & Max prompt length            & 2560 \\
  & Max response length          & 1024 \\
  & Rollouts per prompt ($n$)    & 32 \\
\cmidrule{1-3}
\multirow{5}{*}{Optimization}
  & Learning rate                & $4 \times 10^{-5}$ \\
  & Optimizer                    & AdamW \\
  & $(\beta_1, \beta_2)$         & $(0.9,\ 0.999)$ \\
  & Weight decay                 & 0.01 \\
  & Gradient clip                & 1.0 \\
\cmidrule{1-3}
\multirow{3}{*}{SFT Schedule}
  & Mode                         & interleaved \\
  & SFT steps per cycle          & 50000 \\
  & RL steps per cycle           & 0 \\
\cmidrule{1-3}
\multirow{2}{*}{Infrastructure}
  & Total epochs                 & 100 \\
  & GPU memory utilization       & 0.6 \\
\bottomrule
\end{tabular}
\end{table}

\begin{table}[htbp]
\centering
\caption{Pretraining hyperparameters (OLMo2-1B, 50B tokens).}
\label{tab:hyperparams_pretrain}
\begin{tabular}{lll}
\toprule
\textbf{Category} & \textbf{Hyperparameter} & \textbf{Value} \\
\midrule
\multirow{3}{*}{Data}
  & Total training tokens        & 50B \\
  & Global batch size (sequences) & 512 \\
  & Gradient accumulation steps  & 64 \\
\cmidrule{1-3}
\multirow{5}{*}{Optimization}
  & Learning rate                & $4 \times 10^{-4}$ \\
  & Optimizer                    & AdamW \\
  & $(\beta_1, \beta_2)$         & $(0.9,\ 0.95)$ \\
  & Weight decay                 & 0.1 \\
  & Gradient clip                & 1.0 \\
\cmidrule{1-3}
\multirow{4}{*}{LR Schedule}
  & Schedule                     & cosine with warmup \\
  & Warmup tokens                & 1B \\
  & Min LR ratio ($\alpha_f$)    & 0.1 \\
  & Units                        & tokens \\
\cmidrule{1-3}
\multirow{3}{*}{Regularization}
  & Precision                    & BF16 (AMP) \\
  & Softmax auxiliary loss       & \checkmark \\
  & Auxiliary loss multiplier    & $1 \times 10^{-5}$ \\
\bottomrule
\end{tabular}
\end{table}

\newpage
\section{Additional Results For Section~\ref{sec:main_results}}

\subsection{MATH performance}
\begin{figure}[h]
    \centering
    \includegraphics[width=1.0\linewidth]{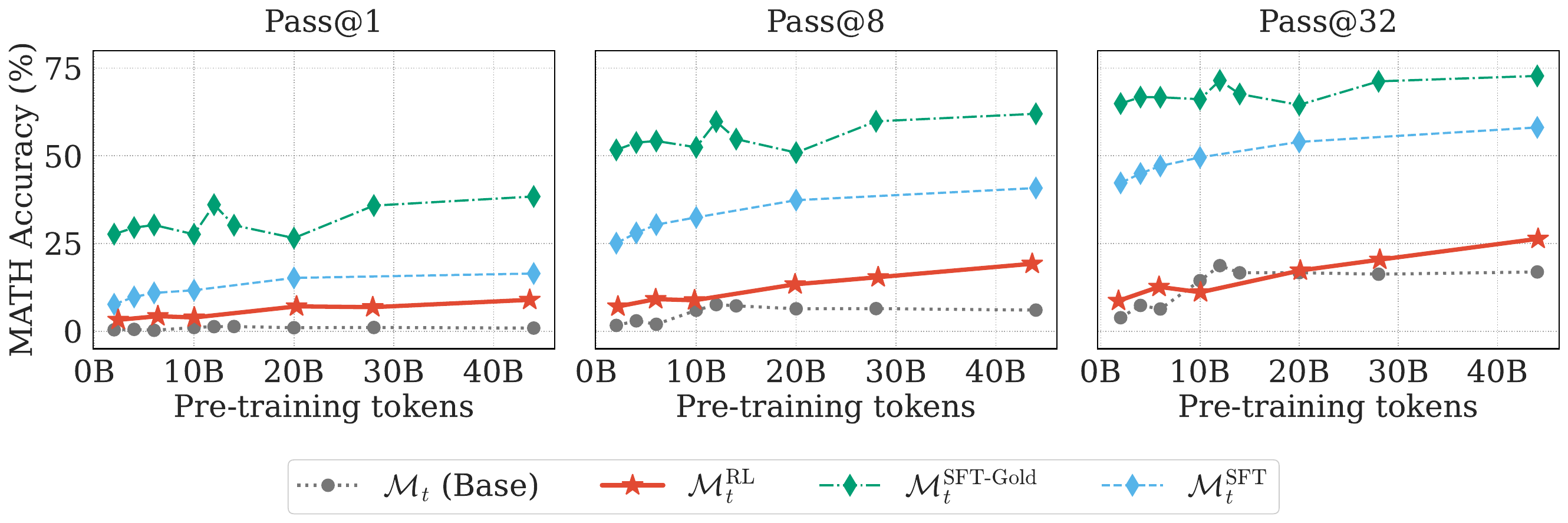}
    \caption{
    \textbf{RL underperforms SFT$\to$RL on harder MATH problems.} MATH $\texttt{pass@k}$ for $\mathcal{M}_t$, $\mathcal{M}_t^{\text{SFT}}$, $\mathcal{M}_t^{\text{SFT} \rightarrow \text{RL}}$, and $\mathcal{M}_t^{\text{RL}}$ trained on the full OpenMathInstruct, with the base model at $N=1$B parameters and $D=50$B pretraining tokens. $\mathcal{M}_t^{\text{RL}}$ still improves over $\mathcal{M}_t$ before Chinchilla-optimal token counts, but a persistent gap to $\mathcal{M}_t^{\text{SFT} \rightarrow \text{RL}}$ remains throughout pretraining, indicating that direct RL is insufficient on harder reasoning tasks.
    }
    \label{fig:omi_pretrain_to_rl_1b}
\end{figure}

\begin{figure}[h]
    \centering
    \includegraphics[width=1.0\linewidth]{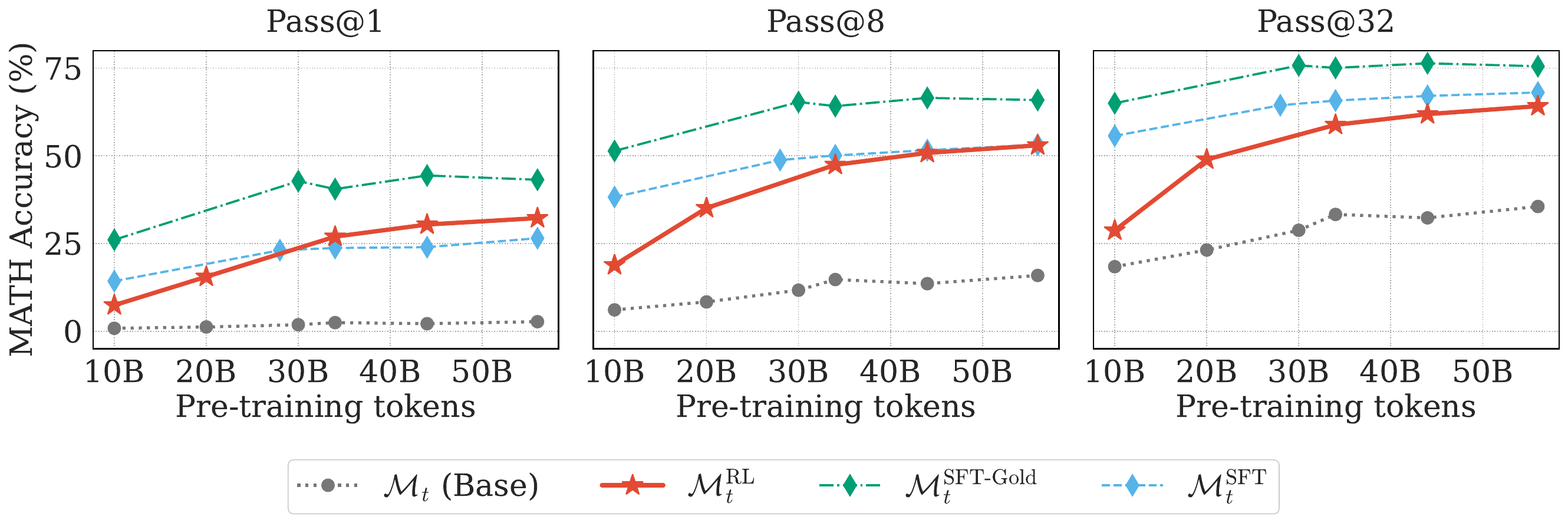}
    \caption{
    \textbf{Adding math pretraining data narrows the MATH gap.} Same setup as \Cref{fig:omi_pretrain_to_rl_1b}, but with 10B additional math-heavy tokens mixed into pretraining ($N=1$B, $D=60$B). Including task-relevant pretraining data substantially boosts $\mathcal{M}_t^{\text{RL}}$ on MATH and narrows the gap to $\mathcal{M}_t^{\text{SFT} \rightarrow \text{RL}}$, supporting pretraining data composition as the binding constraint on early-RL effectiveness.
    }
    \label{fig:omi_pretrain_to_rl_1b_60b}
\end{figure}

\begin{figure}[h]
    \centering
    \includegraphics[width=1.0\linewidth]{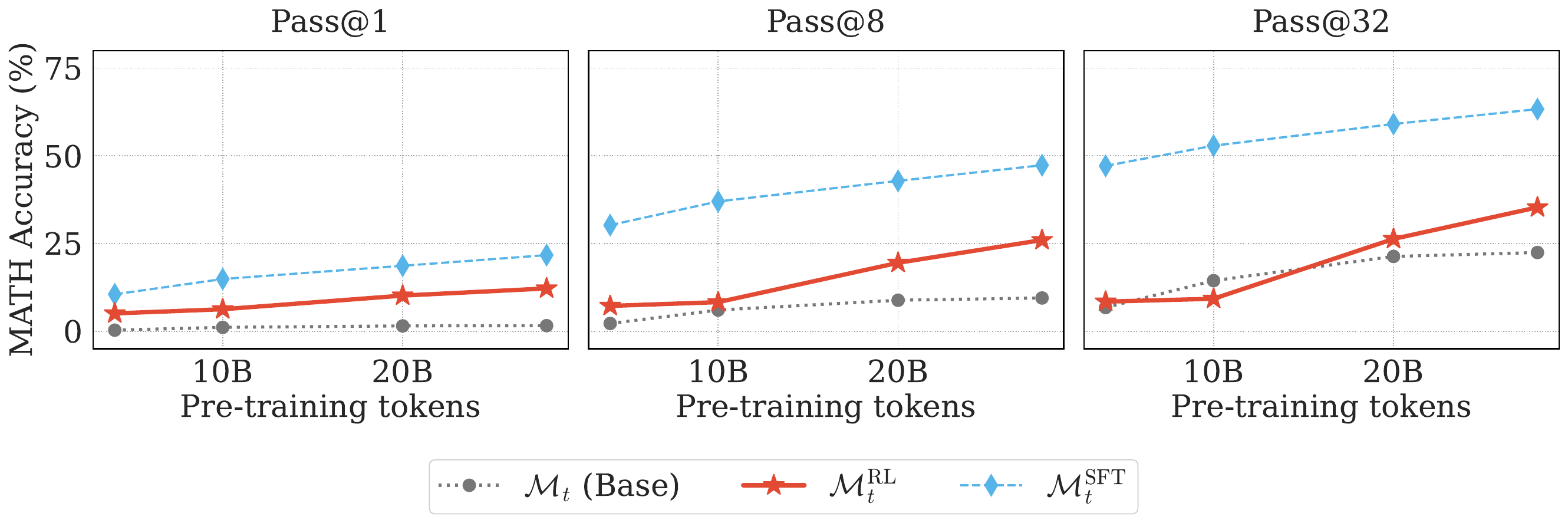}    
    \caption{
    \textbf{Scaling parameters does not close the MATH gap.} Same setup as \Cref{fig:omi_pretrain_to_rl_1b}, but at $N=4$B parameters with the same $D=50$B-token pretraining mix. Increasing model scale improves base-model performance, but does \emph{not} unlock additional RL gains on MATH. In contrast to the data-scaling intervention in \Cref{fig:omi_pretrain_to_rl_1b_60b}, the gap to $\mathcal{M}_t^{\text{SFT} \rightarrow \text{RL}}$ persists.
    }
    \label{fig:omi_pretrain_to_rl_4b}
\end{figure}

See Fig.~\ref{fig:omi_pretrain_to_rl_1b} for MATH performance on original 1B model,  Fig.~\ref{fig:omi_pretrain_to_rl_1b_60b} for MATH performance on 1B model trained on 60B tokens and finally, Fig.~\ref{fig:omi_pretrain_to_rl_4b} for MATH performance on 4B model.

\subsection{Added data for scaling $D$}
\label{app:10b_new_tokens}
We detail the source of the 10B tokens we add into training for the MATH benchmark. 

\begin{table}[t]
\centering
\caption{Composition of the additional math tokens mixed into 
pretraining for the 1B-60B model 
(Section~\ref{subsec:task_difficulty}). All sources are drawn 
from the math subset of the Dolma 3 Dolmino 
Mix~\citep{olmo2025olmo}.}
\label{tab:math_mix}
\begin{tabular}{llr}
\toprule
\textbf{Source} & \textbf{Description} & \textbf{Tokens} \\
\midrule
TinyMATH Mind & Conversational solutions to MATH-style problems & 898M \\
TinyMATH PoT & Program-of-thought solutions to MATH-style problems & 241M \\
CraneMath & Swallow Math reproduction & 5.62B \\
MegaMatt & MegaMath-Web-Pro-Max reproduction & 1.73B \\
\bottomrule
\end{tabular}
\end{table}

\subsection{RL training dynamics}
\label{appdx:gsm_rl_dynamics}
\begin{figure}[h]
    \centering
    \includegraphics[width=1.0\linewidth]{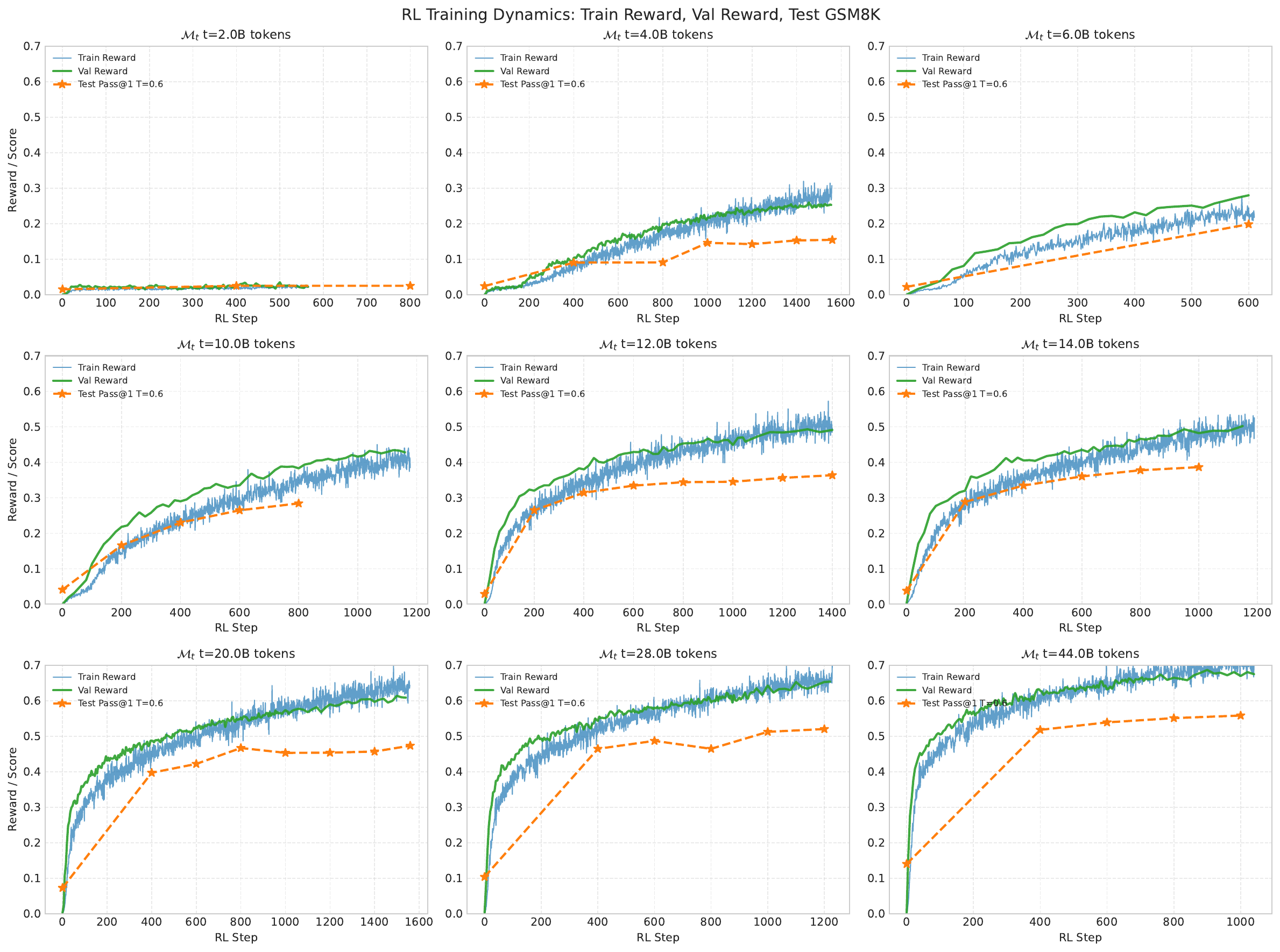}
    \caption{\textbf{RL training reaches convergence at all checkpoints.} Training reward, validation reward, and GSM8K test reward during RL training for $\mathcal{M}_t^{\text{RL}}$ across pretraining checkpoints $t$. All three reward metrics converge by end-of-training, confirming that performance differences between checkpoints are not artifacts of insufficient RL optimization. For checkpoints with seed brittleness ($t < 10$B), we plot the favorable seed; see App.~\ref{appdx:seed_dependency} for seed comparisons.}
    \label{fig:gsm_rl_dynamics}
\end{figure}

In Fig.~\ref{fig:gsm_rl_dynamics}, we show that for all $\mathcal{M}_t^{\text{RL}}$ (across all pretraining checkpoints $\mathcal{M}_t$), the RL training reward, validation reward (computed on a manually split subset of OpenMathInstruct), and GSM8K reward have converged. For earlier checkpoints that exhibit seed brittleness (Sec.~\ref{appdx:seed_dependency}), we report the favorable seed here. See App.~\ref{appdx:seed_dependency} for examples of favorable and unfavorable seeds.

\subsection{Seed dependency}
\label{appdx:seed_dependency}

\begin{figure}
    \centering
    \includegraphics[width=0.9\linewidth]{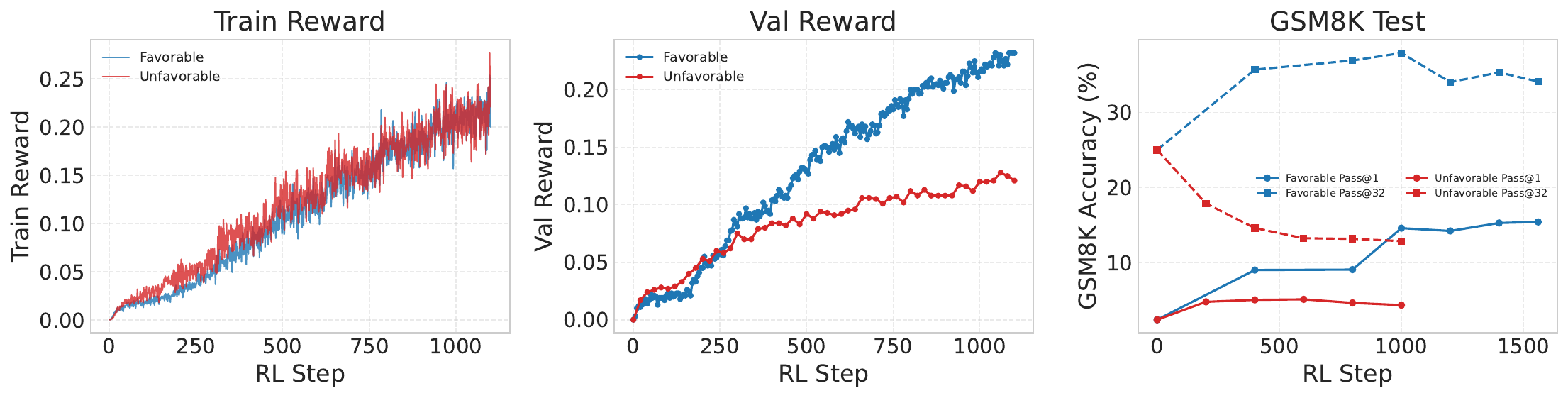}
    \caption{\textbf{Training reward hides RL seed brittleness on early checkpoints.} A favorable seed (\textit{blue}) and an unfavorable seed (\textit{red}) for $\mathcal{M}_t^{\text{RL}}$ at $t=4$B tokens. \textit{Left}: training reward curves are nearly identical between seeds, offering no warning of divergent test outcomes. \textit{Middle}: validation reward begins to diverge mid-training and unfavorable seed only reaches 10\% which comes from format reward. \textit{Right}: on GSM8K, the favorable seed gains substantially on both $\texttt{pass@1}$ and $\texttt{pass@32}$, while the unfavorable seed shows minimal $\texttt{pass@1}$ gain and worsens $\texttt{pass@32}$. This brittleness resolves by $t=10$B tokens.
    \label{fig:seed_dependency}}
\end{figure}
We visualize the outcomes in Figure~\ref{fig:seed_dependency}. Random seed dependency in LLM training has also been observerd in \citet{zhao2026randomscalingemergentcapabilities, Qin2024-jk} as a potential explanation of the emergence phenomenon.

\subsection{SFT dynamics}
\label{appdx:sft_dynamics}
\begin{figure}[ht]
    \centering
    \includegraphics[width=0.8\linewidth]{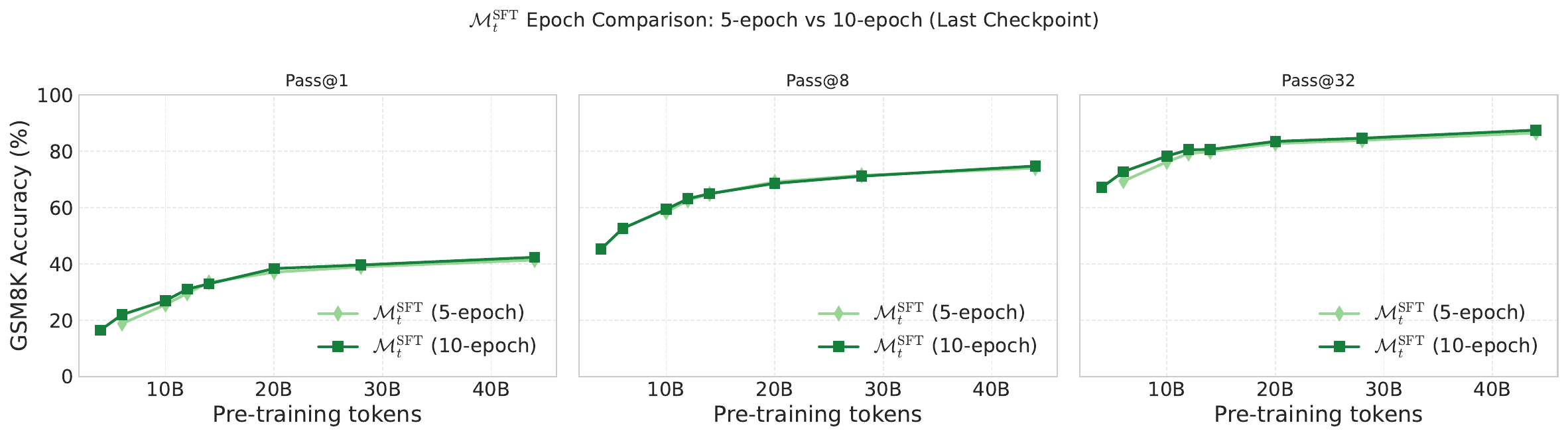}
    \caption{\textbf{SFT converges by 5 epochs.} GSM8K accuracy of $\mathcal{M}_t^{\text{SFT}}$ after training for different numbers of epochs on OpenMathInstruct. Performance plateaus by 5 epochs, which we use as the standard SFT training length for all $\mathcal{M}_t^{\text{SFT}}$ baselines.}
    \label{fig:sft_epochs}
\end{figure}

In Fig.~\ref{fig:sft_epochs}, we experiment with different numbers of SFT epochs to train $\mathcal{M}_t^{\text{SFT}}$ and confirm that 5 epochs leads to convergence in the model's performance. 

\subsection{Evaluating Pretraining Checkpoints}
\label{appdx:n_shot_eval}
\begin{figure}[h]
    \centering
    \includegraphics[width=0.8\linewidth]{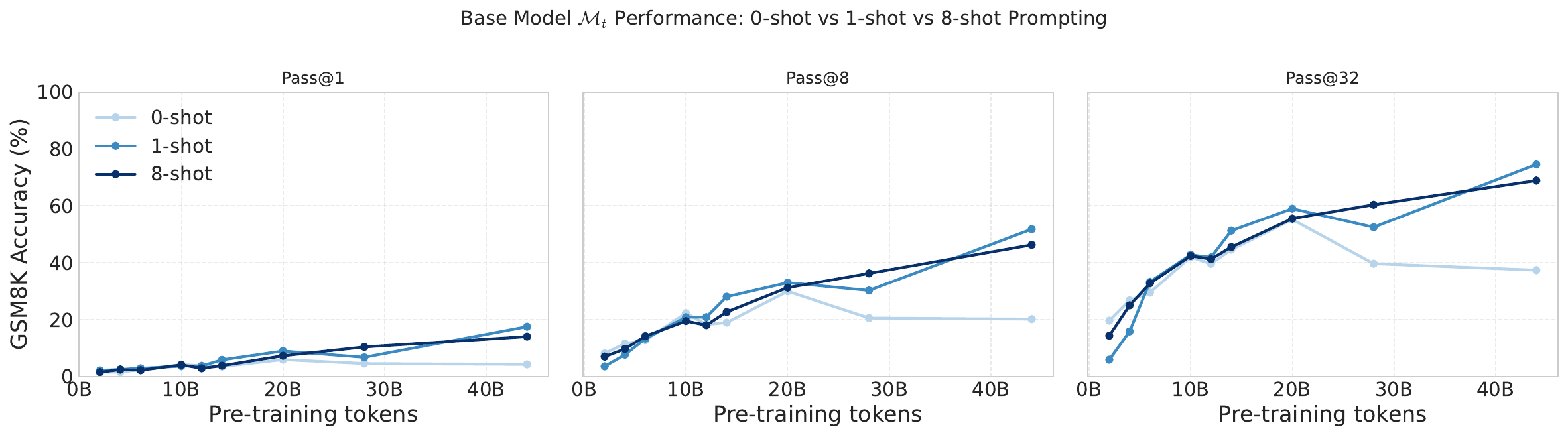}
    \includegraphics[width=0.8\linewidth]{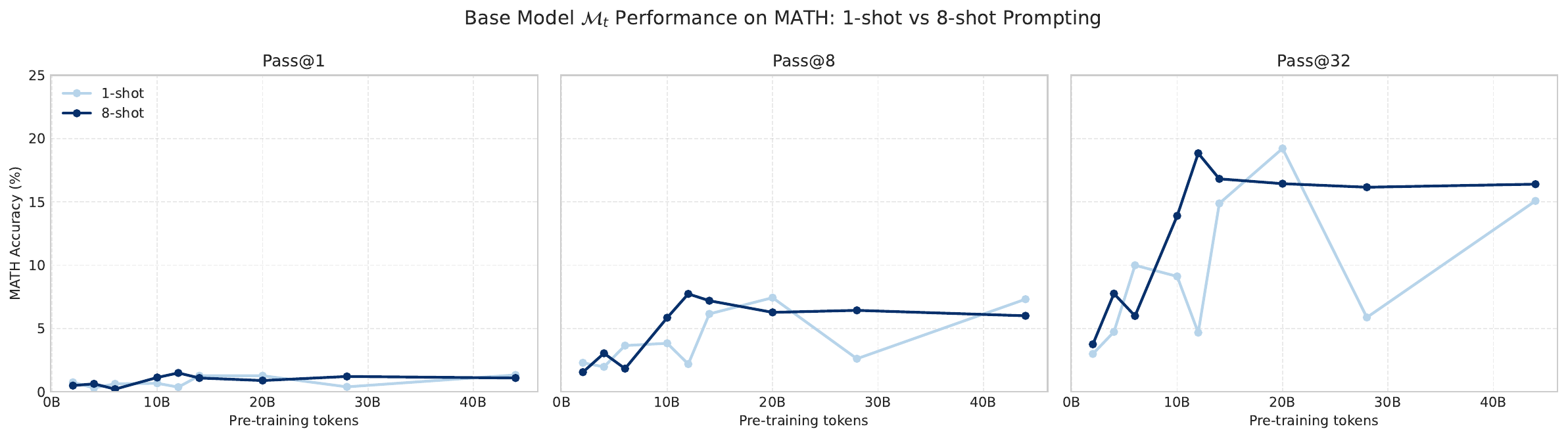}
    \caption{\textbf{Base checkpoints peak at 8-shot prompting.} Performance of pretraining checkpoints $\mathcal{M}_t$ on GSM8K (\textit{top}) and MATH (\textit{bottom}) under varying numbers of in-context examples. Across both benchmarks, accuracy is maximized at 8-shot, which we use throughout for $\mathcal{M}_t$ evaluation.}
    \label{fig:eval_shots}
\end{figure}

In Fig.~\ref{fig:eval_shots}, we experiment with different numbers of in-context examples ($n$-shot) to evaluate the reasoning capabilities of pretraining checkpoints $\mathcal{M}_t$. We confirm that by using 8-shot prompting, the base model achieves the best performance on both MATH and GSM8K.

\section{Full Parallel Average Results}
\begin{figure}
    \centering
    \includegraphics[width=1.0\linewidth]{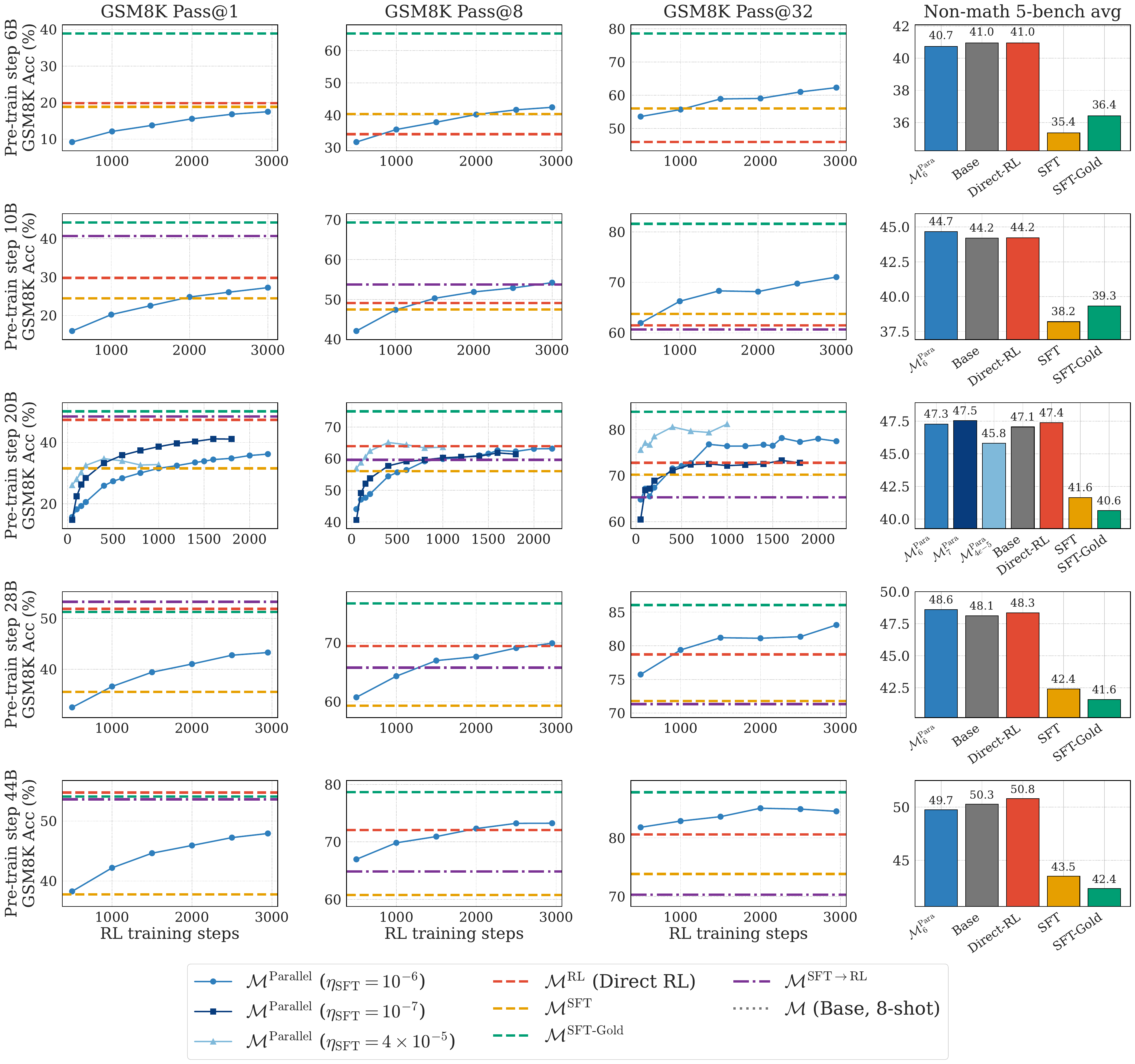}
    \caption{
    \textbf{Full results for parallel average algorithm.}
    }
    \label{fig:paralle_avg_full}
\end{figure}

In \Cref{fig:paralle_avg_full}, we show the full parallel-average training
trajectories at each pre-training checkpoint. \Cref{fig:parallel-avg-main}
summarizes these as a single (final-RL-step) point per checkpoint.

\section{RL Rollouts}
\label{sec:rollouts}

When training with RL on early pretraining checkpoints, the  model is likely to have low $\texttt{pass@k}$ accuracy on the training questions.
Compared to the standard pipeline or a later pretraining checkpoint, applying RL at early pretraining \textit{exacerbates} the reward sparsity problem. On these very early pretraining checkpoints, without sufficient positive samples (i.e., correct rollouts), the learning signal might become sparse or noisy, making it difficult for the model to improve.

A natural strategy to consider in order to obtain higher training signal
is to sample a larger number of rollouts
at each step in training.
In this section, we comprehensively analyze this strategy
and  study the influence of number of rollouts for RL training.
Specifically, we investigate the effect of varying the number of rollouts per prompt ($n$) in GRPO.
We seek to determine if increasing the number of rollouts benefits models that are initially weak on the training distribution.
To this end, we partition our training set into two sets: \textit{a hard set}
 and \textit{an easy set},
simulating early and later stages of pretraining
respectively.
We perform RL using GRPO on both these splits
using settings with few ($n=5$) and many ($n=64$) rollouts and report $\texttt{pass@k}$ accuracy on the standard GSM8K test set.

\subsection{Experimental Setup}
\paragraph{Data and Metrics.}
In order to simulate different stages of pretraining,
we partition our training dataset
based on proportion of positive samples per example. 
The OpenMathInstruct dataset~\citep{toshniwal2024openmathinstruct}
is composed of questions inspired by either MATH or GSM8K training 
sets (for details see,~\S\ref{sec:methods_and_baselines}).
We focus on only the GSM8K-like subset of OpenMathInstruct ($80$K examples).
To define the training splits based on ``difficulty" level,
we evaluate our base model
on the original dataset in a zero-shot setting.
For each question, we generate $64$ responses at temperature $1$ and record the number of correct solutions. We classify questions with $16$ to $64$ correct responses as \textit{GSM8K-Easy}, and those with at most $8$ correct responses as \textit{GSM8K-Hard}.
From these subsets, we randomly sample $10$K questions for each split. We train with GRPO (as described in \S\ref{sec:methods_and_baselines}) and report $\texttt{pass@k}$ ($\texttt{k}\in\{1, 8\}$) metrics on the standard GSM8K test set.

\begin{figure}[t]
  \centering
  \begin{minipage}{0.95\textwidth}
    \centering
    \includegraphics[width=\linewidth]{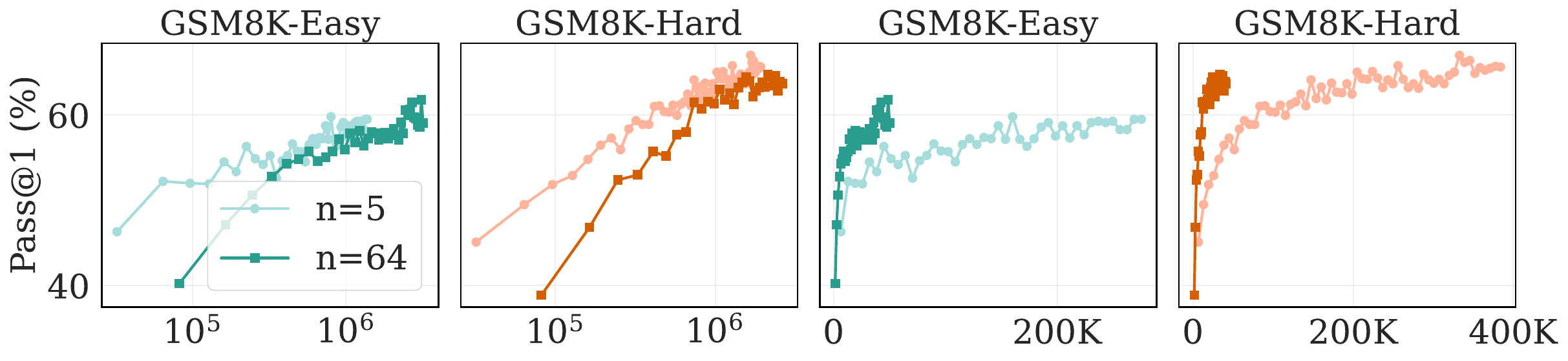}
  \end{minipage} \vfill
  \begin{minipage}{0.95\textwidth}
    \centering
    \includegraphics[width=\linewidth]{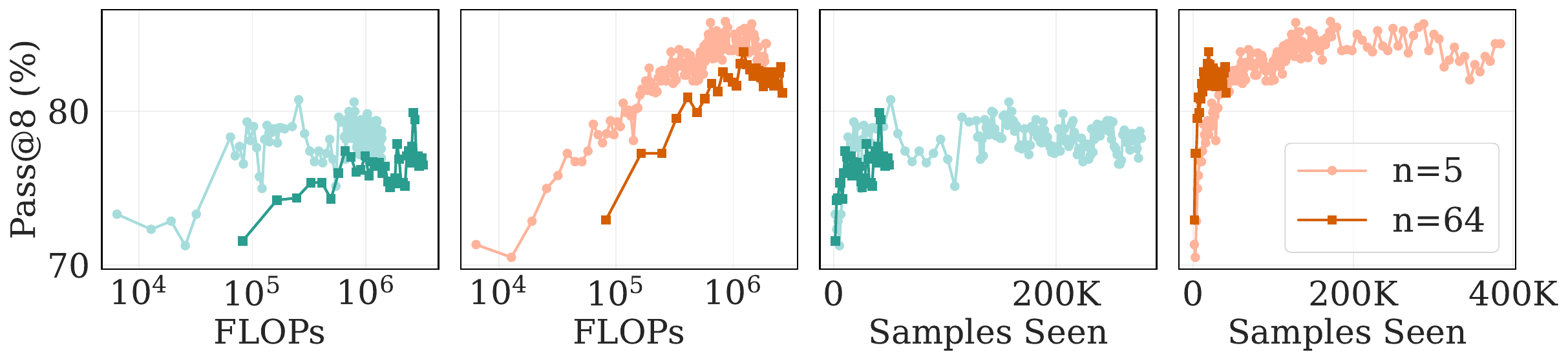}
  \end{minipage}
  \caption{
  \label{fig:gsm8k_rollouts}
  \textbf{Fewer rollouts are more FLOP-efficient at convergence.} GSM8K $\texttt{pass@k}$ during RL training with $n=5$ versus $n=64$ rollouts per prompt, on training sets sub-sampled to be relatively easy or hard for the base model (a proxy for late vs.\ early pretraining). Asymptotic performance is similar across rollout counts. However, $n=5$ achieves comparable accuracy at substantially lower FLOPs, especially on the harder split. Larger $n$ is more sample-efficient per training example, but not per FLOP.
  }
\end{figure}

\paragraph{Model and Method.} 
We conduct all experiments using the OLMo2 1B model~\citep{olmo20242}.

We perform GRPO training for both GSM8K-Easy and GSM8K-Hard using $n=5$ and $n=64$ rollouts per prompt, while keeping all other hyperparameters constant.
Consequently, $n=64$ consumes significantly more FLOPs per RL step. To account for this trade-off, we analyze accuracy as a function of both total FLOPs consumed and the number of examples during RL training. For all settings, we train the models until the validation \texttt{pass@1} metric converges.

\subsection{Main Results } 
We observe a distinct trade-off between sample efficiency and compute efficiency. As a function of \textit{samples seen}, increasing the number of rollouts to $n=64$ greatly improves $\texttt{pass@1}$ convergence compared to $n=5$. However, when viewed as a function of \textit{FLOPs}, the lower rollout setting ($n=5$) is more compute-efficient in the early stages of training. As training progresses toward $10^6$ FLOPs, this efficiency gap narrows, with $n=64$ eventually matching or surpassing the performance of $n=5$.
%
We observe that the difference between $n=5$ and $n=64$ rollouts further diminishes when observing $\texttt{pass@1}$. However, when we match FLOPs, we see that $n=5$ appears to significantly improve upon $n=64$, especially when training with GSM8K-Hard. 

%
Our analysis in \Cref{fig:gsm8k_rollouts} yields three primary insights regarding the scaling of RL rollouts. \textbf{First}, we find that asymptotic performance is largely independent of the number of rollouts; both $n=5$ and $n=64$ converge to similar $\texttt{pass@k}$ peaks across difficulty levels. \textbf{Second}, there is a clear trade-off between sample efficiency and compute efficiency. Increasing the rollout count ($n=64$) maximizes the utility of each training example, leading to faster convergence in terms of training steps. Conversely, reducing the rollout count ($n=5$) is significantly more FLOP-efficient, achieving comparable performance with a fraction of the compute budget. \textbf{Finally}, this compute advantage is particularly pronounced on the \textit{GSM8K-Hard} split for the $\texttt{pass@8}$ metric, suggesting that when rewards are sparse (as with early checkpoints), massive rollout scaling may yield diminishing returns per FLOP compared to processing more batches with fewer rollouts.


\end{document}